\documentclass[sigconf]{aamas} 
\usepackage{balance} 
\usepackage{hyperref}       
\usepackage{url}            
\usepackage{booktabs}       
\usepackage{amsfonts}       
\usepackage{nicefrac}       
\usepackage{microtype}      
\usepackage{xcolor}         
\usepackage{caption}
\usepackage{subcaption}
\usepackage[export]{adjustbox}
\usepackage{float}
\usepackage{geometry}
\setcitestyle{square,numbers}
\usepackage{mathtools}
\usepackage{amsthm}
\theoremstyle{plain}
\newtheorem{theorem}{Theorem}[section]

\theoremstyle{definition}
\newtheorem{definition}[theorem]{Definition}

\theoremstyle{remark}

\DeclareMathOperator*{\argmax}{arg\,max}

\newcommand{\klprior}{p^{\text{prior}}_{\theta}}
\usepackage[inkscapeformat=png]{svg}
\usepackage{comment}

\def\BibTeX{{\rm B\kern-.05em{\sc i\kern-.025em b}\kern-.08em
    T\kern-.1667em\lower.7ex\hbox{E}\kern-.125emX}}

\setcopyright{ifaamas}
\acmConference[AAMAS '26]{Proc.\@ of the 25th International Conference
on Autonomous Agents and Multiagent Systems (AAMAS 2026)}{May 25 -- 29, 2026}
{Paphos, Cyprus}{C.~Amato, L.~Dennis, V.~Mascardi, J.~Thangarajah (eds.)}
\copyrightyear{2026}
\acmYear{2026}
\acmDOI{}
\acmPrice{}
\acmISBN{}
\acmSubmissionID{71}
\title{Multi-Agent Model-Based Reinforcement Learning with Joint State-Action Learned Embeddings}

\author{Zhizun Wang}
\affiliation{
  \institution{McGill University}
  \city{Montreal}
  \country{Canada}
  }
\email{zhizun.wang@mail.mcgill.ca}
\author{David Meger}
\affiliation{
  \institution{McGill University}
  \city{Montreal}
  \country{Canada}
  }
\email{david.meger@mcgill.ca}

\begin{abstract}
Learning to coordinate many agents in partially observable and highly dynamic environments requires both informative representations and data-efficient training. To address this challenge, we present a novel model‑based multi‑agent reinforcement learning framework that unifies joint state-action representation learning with imaginative roll-outs. We design a world model trained with variational auto‑encoders and augment the model using the state‑action learned embedding (SALE). SALE is injected into both the imagination module that forecasts plausible future roll-outs and the joint agent network whose individual action values are combined through a mixing network to estimate the joint action-value function. By coupling imagined trajectories with SALE‑based action values, the agents acquire a richer understanding of how their choices influence collective outcomes, leading to improved long‑term planning and optimization under limited real‑environment interactions. Empirical studies on well-established multi‑agent benchmarks, including StarCraft II Micro-Management, Multi‑Agent MuJoCo, and Level‑Based Foraging challenges, demonstrate consistent gains of our method over baseline algorithms and highlight the effectiveness of joint state-action learned embeddings within a multi-agent model‑based paradigm.
\end{abstract}


\begin{document}

\pagestyle{fancy}
\fancyhead{}
\maketitle 

\section{Introduction}

Reinforcement learning (RL) has been established as a fundamental approach for sequential decision-making in complex and uncertain environments. In particular, multi-agent reinforcement learning (MARL) extends RL to scenarios where multiple autonomous agents interact with each other in a shared environment, often requiring collaboration to accomplish a global objective \cite{zhang_multi-agent_2021, Zhu2022ASO}. 
Due to partial observability, scalability, and non-stationarity issues posed by multi-agent systems \cite{gronauer_multi-agent_2021, YangDBLP:journals/corr/abs-2011-00583}, model-free MARL algorithms may struggle to achieve sample efficiency and generalization ability \cite{tampuu_multiagent_2017, DBLP:journals/corr/LoweWTHAM17, rashid_qmix_2018, Iqbal2018ActorAttentionCriticFM}.
One promising solution to sample inefficiency in single-agent RL is feature learning \cite{fujimoto2022should, fujimoto2024sale}. \cite{fujimoto2024sale} proposes \textit{state-action learned embeddings} (SALE), learning a predictive embedding space that captures how actions causally affect the states. SALE jointly constructs state and action representations, effectively learning the transition dynamics in a compact latent space. This allows for more structured representations and opens the door to imagination-based planning. As an alternative solution, model-based reinforcement learning (MBRL) has also demonstrated sample efficiency in solving complex single-agent tasks \cite{pmlr-v97-hafner19a, Hafner2020Dream, NEURIPS2022_Hafner, wu2023models, NEURIPS2019_Janner}, since it requires a much smaller number of samples for training compared to model-free RL \cite{luo2024survey}.

In this paper, we leverage joint state-action representation learning in model-based RL to address the sample complexity problem in multi-agent systems. More specifically, we incorporate SALE into a world model that learns the dynamics of the real environment and generates informative latent space roll-outs. We then present Multi-Agent Model-Based Framework with Joint State-Action Learned Embeddings (MMSA), a novel MARL framework in which the world model works as an imagination module. The predictions of the world model are learned by variational auto-encoders (VAEs). In this framework, we also propose applying SALE to the joint policy network and the joint agent network representing all agents in the environment. Aggregating the world model roll-outs with the global state and the individual action values from the agent network, we pass them into a mixing network \cite{rashid2020monotonic} that employs value factorization for approximating joint action-value functions. Our framework benefits from SALE because it can capture the underlying dynamics between states and actions. It efficiently extracts meaningful features from the latent dynamics, leading to more effective and stable model learning. Moreover, our method uses the mixing network following the paradigm of centralized training with decentralized execution (CTDE) \cite{Kraemer2016MultiagentRL}, ensuring efficient coordination among multiple agents in complex systems \cite{rashid2020monotonic}.

The key contributions of our paper are summarized as follows:

\textbullet \textbf{ Dynamics model for sample efficiency. }
We fuse two effective methods for reducing sample complexity, namely model-based RL and representation learning, by constructing a world model equipped with SALE. Reinforced by joint state-action learned features, the model can supply agents with structured and informative latent representations that markedly improve sample efficiency. 

\textbullet \textbf{ Unified MARL framework with imagination-based value decomposition. }
SALE was initially proposed to improve TD7 \cite{fujimoto2024sale}, and the mixing network was originally part of QMIX \cite{rashid_qmix_2018}, but both TD7 and QMIX are model-free RL algorithms. In contrast, we propose MMSA, a model-based framework that uses a world model to augment the mixing network with simulated trajectories. Each individual agent in the joint agent network is also enhanced with SALE. This means the individual action-value function considers the learned representations from SALE in addition to the local action-observation history. The mixing network aggregates individual action values into a global estimate, preserving the CTDE paradigm.

\textbullet \textbf{ Comprehensive benchmarking and ablation studies. }
We evaluate MMSA on three widely used test beds: Multi-Agent MuJoCo \cite{Peng2020FACMACFM}, StarCraft II MARL benchmark \cite{samvelyan2019, ellis2023smacv2}, and Level-Based
Foraging \cite{Christianos10.5555/3495724.3496622}. Across a rich variety of environments, we have constantly observed the performance of MMSA matching or exceeding the competitive baselines. Moreover, we conduct in-depth design studies to explore the optimal design choices and present systematic ablations to demonstrate that every architectural ingredient contributes meaningfully to the overall performance gains.

\begin{figure}[t]
\centering
\includegraphics[width=\columnwidth]{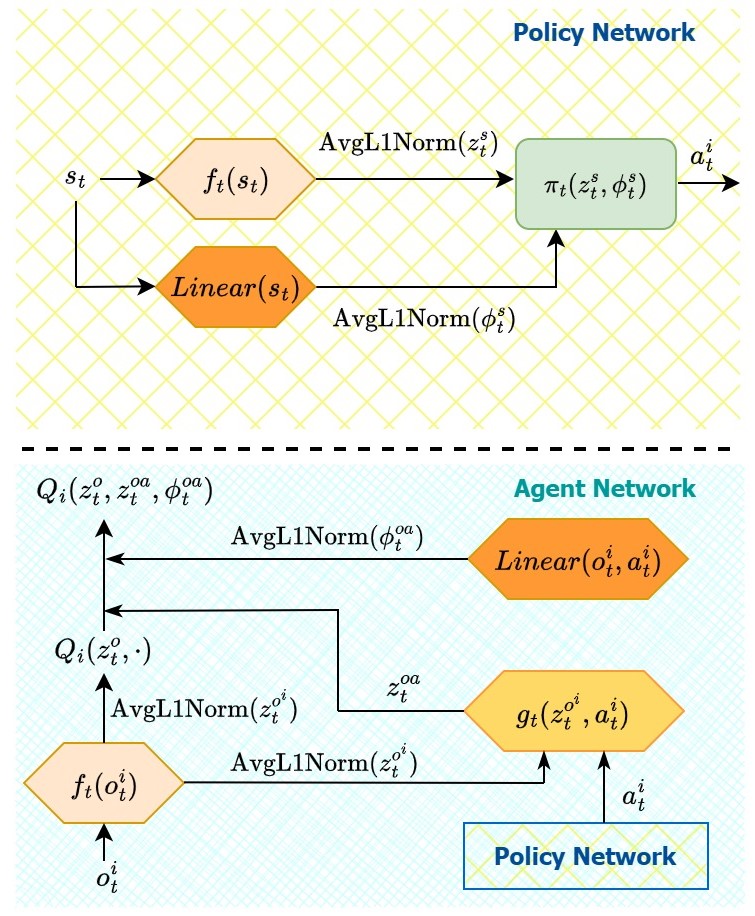}
\caption{Architecture of the SALE-augmented policy and agent networks in MMSA. \textbf{Top:} the policy network in which the state $s_t$ is encoded and passed into $\pi_t$ to produce the action. \textbf{Bottom:} the agent network, which encodes the observation and action for computing $Q_i\bigl( z^o_t,\; z^{oa}_t,\;\phi^{oa}_t\bigr).$}
\label{fig:agent}
\Description{Architecture of the SALE-augmented policy network and agent network in MMSA.}
\end{figure}

\begin{figure*}[htbp]
\centering
\includegraphics[width=\linewidth]{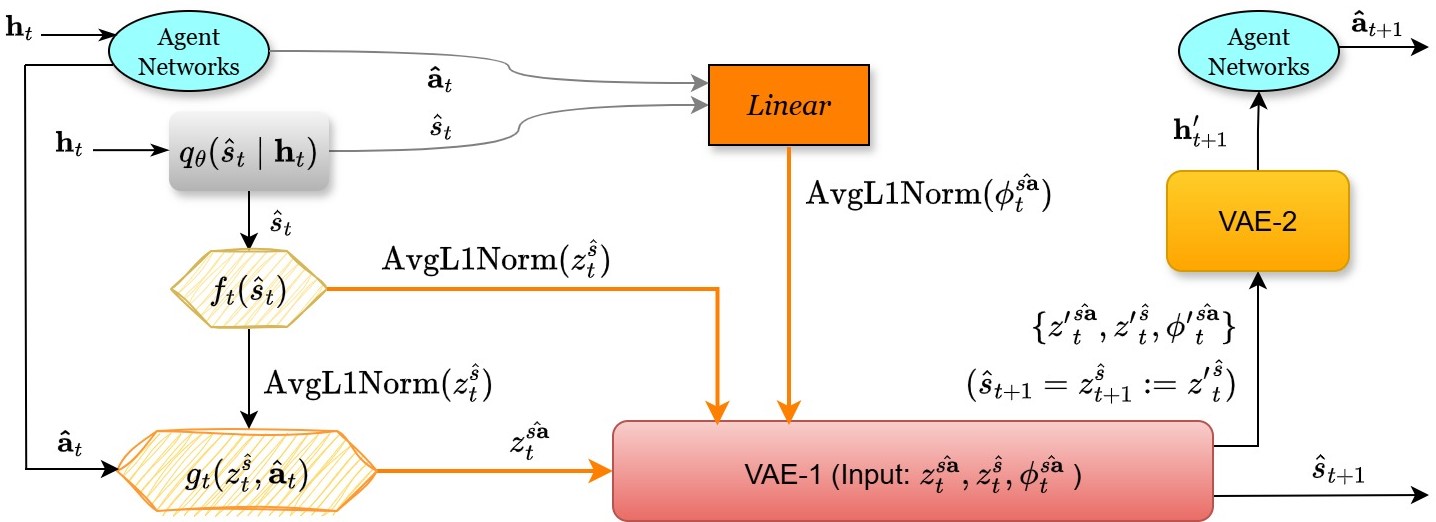}
\caption{Illustration of the world model imagination in MMSA. The input $\mathbf{h}_{t}$ encapsulates the past information, including $\hat{s}_{t-1}$ and $\mathbf{a}_{t-1}$. The agent networks receive $\mathbf{h}_{t}$ and infer $\hat{\mathbf{a}}_{t}$. Taking the normalized joint state-action learned embeddings $({z}_{t}^{\hat{s\mathbf{a}}}, z_{t}^{\hat{s}}, {\phi}_{t}^{\hat{s\mathbf{a}}})$ as input, VAE-1 reconstructs ${z'}_{t}^{\hat{s\mathbf{a}}}, {z'}_{t}^{\hat{s}},$ and $ {\phi'}_{t}^{\hat{s\mathbf{a}}} $. The outputs are passed into VAE-2 to infer $\mathbf{h'}_{t+1}$.}
\label{fig:wmodel}
\Description{Demonstration of the world model imagination in MMSA.}
\end{figure*}

\section{Related Work}


\subsubsection*{Model-Based Reinforcement Learning (MBRL)}
MBRL represents a significant paradigm in reinforcement learning (RL), distinguished by its utilization of an explicit model of the environment \cite{sutton1991dyna, luo2024survey, Deisenroth2011PILCOAM}. The model refers to the abstraction of the environment dynamics, which can be formulated as a Markov decision process (MDP) \cite{Sutton10.5555/3312046, silver_reward_2021}. Analogous to humans' ability to imagine what may happen in the future if different actions are taken, the model can work as a simulation module that helps to choose the correct actions from imagination \cite{pmlr-v97-hafner19a, wu2023models}. Unlike model-free approaches, which rely solely on direct interactions with the environment and can be sample inefficient, MBRL predicts future transitions and rewards, enhancing learning efficiency and decision-making \cite{gu2016continuous, Chua2018DeepRL, Xu2018AlgorithmicFF, NEURIPS2019_Janner, Wang/corr/abs-1907-02057}. 
Built on the foundation of MBRL, the world model \cite{NEURIPS2018_Ha, Ha2018WorldM} assists the agents with accurate knowledge representations and effective behavior learning. This line of work evolves into recurrent state‑space models (RSSM) \cite{pmlr-v97-hafner19a} and the Dreamer family \cite{Hafner2020Dream, hafner2021mastering, NEURIPS2022_Hafner, Hafner2023MasteringDD}, highlighting the power of coupling representation learning, predictive modeling, and policy optimization within a single algorithmic framework. 
Recent efforts tackle limitations of MBRL frameworks and explore hybrid designs \cite{ NEURIPS2020_Shen, chebotar_actionable_2021, Shen3495724.3495961, Luo2020PMADRLAP}. \cite{racaniere2017imagination} combines a model‑free policy path with a rollout encoder that takes simulated trajectories from a model-based path and employs a policy module to determine the imagination-augmented policy. \cite{NEURIPS2022_Pan} separates controllable from uncontrollable factors in visual control tasks, improving robustness in non‑stationary tasks.

\subsubsection*{Multi-Agent Reinforcement Learning (MARL)}
MARL studies how multiple learners interact in a shared environment, where the agents are confronted with non‑stationary dynamics, partial observability, and combinatorial complexities \cite{Claus1998TheDO, 10.1145/860575.860689, gronauer_multi-agent_2021, Zhu2022ASO, YangDBLP:journals/corr/abs-2011-00583}. Analogous to the single-agent domain, two principal MARL algorithmic classes have emerged: value-based methods, which compute value function estimates of the agents \cite{DBLP:journals/corr/abs-2006-00587, Oliehoek2008OptimalAA, liu2023CIA/ALL}, and policy gradient methods, which update the learning parameters along the direction of the gradient of specific metrics with respect to the policy parameter \cite{Sutton10.5555/3312046, zhang_multi-agent_2021, foerster_counterfactual_2018, Peng2020FACMACFM, Iqbal2018ActorAttentionCriticFM, Zhou2020LearningIC, Christianos10.5555/3495724.3496622, DBLP:journals/corr/LoweWTHAM17}. Value-based MARL methods range from complete decentralization \cite{Tan1997MultiAgentRL, tampuu_multiagent_2017} to full centralization \cite{Boutilier10.5555/1029693.1029710, Guestrin2002CoordinatedRL}. Recent studies focused on MARL algorithms that lie between the two extremes of centralization \cite{son_qtran_2019, rashid2020monotonic, tabishDBLP:journals/corr/abs-2006-10800}, according to the CTDE paradigm \cite{Kraemer2016MultiagentRL}. Value decomposition network (VDN) \cite{sunehag_value-decomposition_2017} constructs the joint value function $Q_{tot}$ of the learning agents, which can be additively factorized into individual Q-functions $\tilde{Q}_{i}$. QMIX \cite{rashid_qmix_2018} replaces the full factorization in VDN with the enforcement of monotonicity between $Q_{tot}$ and individual $\tilde{Q}_{i}$, which enables it to represent a larger class of Q-functions than VDN.

Model-based MARL (or equivalently, multi-agent MBRL) is an interesting and emerging discipline with huge potential in real-world applications \cite{YangDBLP:journals/corr/abs-2011-00583, luo2024survey, wang_model-based_2022, egorov2022scalable, sessa2022efficient}. Early development of multi-agent MBRL has mainly focused on theoretical analyses \cite{Brafman29262f, brafman_r-max--general_2003}, and existing algorithms may rely heavily on specific prior knowledge, such as global states and opponent policies \cite{Bai2020ProvableSA, Park2019SymmetricGC, zhang_multi-agent_2021}. \cite{ma2024efficient} devises model-based decentralized policy optimization, reducing reliance on global communication while maintaining performance. \cite{ma2024efficient} focuses on large-scale network control problems that involve traffic networks, whereas our application lies in the areas of multi-agent gaming and robotic control. \cite{pasztor2021efficient} proposes $\text{M}^3$-UCRL, integrating mean‑field game theory with model‑based exploration to obtain sub‑linear regret bounds. However, its reliance on the mean‑field approximation restricts its usage to large systems with many agents only. \cite{ZhangNEURIPS2020_0cc6ee01} conducts a rigorous theoretical analysis on model-based MARL but only focuses on zero-sum Markov games. \cite{zhang_multi-agent_2021} introduces a framework to improve the efficiency of multi-agent policy optimization in competitive settings. However, the framework requires prior knowledge about opponents and therefore risks sizeable generalization errors when the assumptions break.

\subsubsection*{Representation Learning}
Early studies framed representation learning as state abstraction, where bisimulation metrics or MDP homomorphisms are used to collapse an MDP into a smaller decision process \cite{LiWL06, Ferns2011, Chua2018DeepRL}. For high-dimensional spaces, representation learning embeds observation data, such as images, into compact latent vectors for control \cite{Watter2015, Finn2016, mnih_human-level_2015, Lillicrap2015ContinuousCW}. Another mainstream interpretation is feature learning through auxiliary signals, which shape latent spaces towards the aspects of the environment most relevant for decision‑making \cite{conf/atal/SuttonMDDPWP11, Jaderberg2017, pmlr-v80-riedmiller18a}. As a newly developed representation learning method built on OFENet \cite{Ota2020}, SALE \cite{fujimoto2024sale} demonstrates the importance of learning low-level state-action representations in understanding the complexity of dynamical systems.

\section{Background}

\subsubsection*{Dec-POMDP}
The decentralized partially observable Markov decision process (Dec-POMDP) is appropriate for modeling collaborative agents in partially observable scenarios \cite{oliehoek2016concise}. 
A Dec-POMDP is defined by a tuple 
$ M := \langle \mathcal{S}, \mathcal{A}, \mathcal{N}, T, \Omega, O, R, \gamma \rangle, $
where $\mathcal{S}$ is the state space of all agents, $\mathcal{A}$ is the joint action space of all agents, $\mathcal{N} := \{ 1,..., N \}$ represents the set of $N$ agents, $T$ is the state transition function, $\Omega$ represents the observation space, $O$ is the observation function, $R$ is the reward function, and $\mathit{\gamma} \in [0,1]$ represents the discount factor with respect to time.
At time step $t$, each agent $i \in \mathcal{N}$ chooses an action $a^i$ from its own action space $\mathcal{A}^i$ to form the joint action $\mathbf{a} \in \mathcal{A}$, where $\mathbf{a} := (a^1,...,a^N)$ and $\mathcal{A} := \times_{i \in \mathcal{N}} \mathcal{A}^i$. Then the environment moves from $s$ to $s'$ based on $T(s' | s, \mathbf{a}): \mathcal{S} \times \mathcal{A} \times \mathcal{S} \to [0, 1].$ Every agent $i$ draws an observation $o \in \Omega$ according to $O(s, i): \mathcal{S} \times \mathcal{N} \to \Omega$ because of partial observability. $i$ has its own action-observation history, denoted by $\tau^i \in \mathcal{T}^i := (\Omega \times \mathcal{A}^i)^{*}$, and selects $a^i$ by its policy $\pi^i(a^i | \tau^i ): \mathcal{T}^i \times \mathcal{A}^i \to [0, 1].$ 
The learning goal is to maximize the expected return by optimizing the joint policy ${\pi} := (\pi^1, ..., \pi^N)$. The joint action-value function of ${\pi}$ is written as

$\phantom{q-function} Q^{\pi}(s_t, \mathbf{a}_t) = \mathbb{E}_{s_{t+1:\infty}, \mathbf{a}_{t+1:\infty}}[G_t | s_t, \mathbf{a}_t], $

where $G_t = \sum^{\infty}_{k=0} \gamma^k r_{t+k}$ is the discounted return; $r_{t+k}$ is the reward computed by $R$ for all agents at time step $t+k$.

\section{Method}

We propose a multi-agent MBRL approach named Multi-Agent Model-Based Framework with Joint State-Action Learned Embeddings (MMSA). Our method couples a world model augmented by state-action representation learning and a value decomposition framework under the CTDE paradigm. We begin by formalizing the amortized variational inference problem in Dec-POMDPs, deriving the corresponding evidence lower bound (ELBO), and analyzing the optimization process for the ELBO. In the following subsection, we introduce two key components for MMSA: a policy network that maps state embeddings to actions and an agent value network that evaluates local action-value functions before they are combined by a monotonic mixing network.
We then elaborate on the world model, which employs SALE encoders to generate roll‑outs in latent space, providing synthetic experience without additional environment interaction. The details of the full MMSA workflow are presented next, showing how these modules interact under the CTDE paradigm. The section concludes with the unified learning objective, which involves loss functions for KL regularization, VAE reconstruction, temporal‑difference (TD), and SALE prediction.

\begin{figure*}[htbp]
\centering
\includegraphics[width=\linewidth]{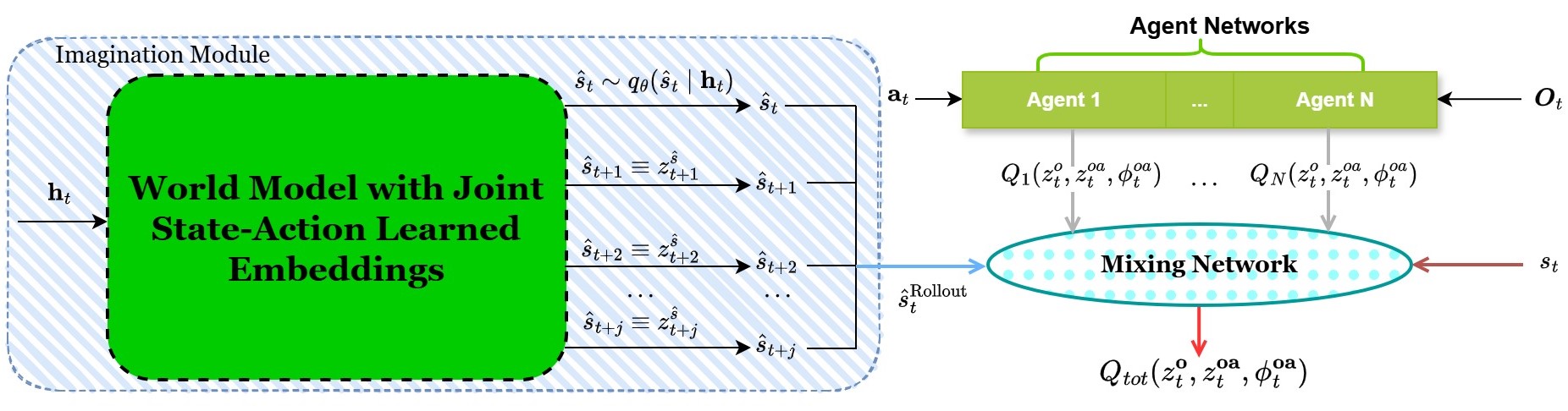}
\caption{An overview of the MMSA method that illustrates how our model-based MARL framework weaves together (1) a learned world model with state-action learned embeddings, (2) decentralized agent value networks equipped with SALE, and (3) a QMIX-style mixing network under the CTDE paradigm. The learning process of the world model is shown in Figure \ref{fig:wmodel}. }
\label{fig:pipeline}
\Description{An overview of the MMSA framework.}
\end{figure*}

\subsection{Deriving and Optimizing the Variational Lower Bound}
\label{sec4.1}

To model the dynamics of multi-agent systems and perform model learning, we need to analyze how the evidence lower bound (ELBO) for the latent state inference in the world model can be deduced and optimized when the system is modeled as a Dec-POMDP.
We first denote $\hat{s}_t$ as an abstraction of the agents' local observations $\mathbf{o}_{t}$, which implies $\mathbf{o}_{t} \sim p(\mathbf{o}_{t} | \hat{s}_t)$ by definition. Inspired by \cite{Huang2020SVQNSV}, we introduce two approximators $q_{\theta}(\cdot)$ and $q_{\pi}(\cdot)$. The function $q_{\pi}(\cdot)$ approximates the optimal policy, and $q_{\theta}(\cdot)$ is the inference function for the latent state space, where $\theta$ represents the learnable parameters. When $q_{\theta}(\cdot)$ is fixed, $q_{\pi}(\cdot)$ can be trained using vanilla Q-learning. When $q_{\pi}(\cdot)$ is fixed as the optimal policy, $q_{\theta}(\cdot)$ can be learned. Consider the joint observation $\mathbf{o} \in \mathcal{O}$ and joint action $\mathbf{a} \in \mathcal{A}$. We can define the approximate posterior as 
$q_{\theta}(\hat{s}_t | \hat{s}_{t-1}, \mathbf{a}_{t-1}, \mathbf{o}_{t} )$ at time step $t \in [0, T]$. This function is used to infer the latent states. We can then derive the ELBO of Dec-POMDP as follows.
\begingroup
\allowdisplaybreaks
\begin{align}
&\mathcal{L}_{\text{ELBO}} (\mathbf{a}_{0:T}, \mathbf{o}_{1:T}) 
= \log p(\mathbf{a}_{0:T}, \mathbf{o}_{1:T}) \nonumber \\
&= \log \mathbb{E}_{q_{\theta}(\hat{s}_{1:T} | \mathbf{a}_{0:T}, \mathbf{o}_{1:T} )} 
\left[ \frac{p(\hat{s}_{1:T}, \mathbf{a}_{0:T}, \mathbf{o}_{1:T} )}{q_{\theta}(\hat{s}_{1:T} | \mathbf{a}_{0:T}, \mathbf{o}_{1:T} )} \right]  \nonumber \\
& \approx \sum_{t=1}^{T} \{ \log \left[ p(\mathbf{a}_{t} | \mathbf{o}_{t}) \right] + 
\log \left[ p(\mathbf{o}_{t} | \hat{s}_{t}) \right]  
- \nonumber \\
& \phantom{=} \mathcal{D}_{\text{KL}} \left[ q_{\theta}(\hat{s}_t | \hat{s}_{t-1}, \mathbf{a}_{t-1}, \mathbf{o}_{t} ) \parallel
p(\hat{s}_t | \hat{s}_{t-1}, \mathbf{a}_{t-1}) \right] \} 
\label{mtd1} 
\end{align}
\endgroup

We derive $\mathcal{L}_{\text{ELBO}}$ with respect to the joint action and joint observation to solve the amortized variational inference problem in MARL. In contrast, \cite{Huang2020SVQNSV} only studies the POMDP in single-agent domains.
The detailed deduction steps can be found in Appendix \ref{appx:elbo} in the supplementary material.

The term $\log \left[ p(\mathbf{a}_{t} | \mathbf{o}_{t}) \right]$ in Eq.\ref{mtd1} represents the joint policy, which is irrelevant to the state inference problem. The last term of $\mathcal{L}_{\text{ELBO}}$ denotes the negative Kullback-Leibler (KL) divergence, which implies that the KL distance between the posterior and prior approximates should be minimized. Since the actual prior distribution $p(\hat{s}_t | \hat{s}_{t-1}, \mathbf{a}_{t-1})$ is unknown, we introduce a generative model $\klprior(\hat{s}_t | \hat{s}_{t-1}, \mathbf{a}_{t-1})$ to estimate the prior. 
We use variational auto-encoders (VAEs) to learn $\klprior(\hat{s}_t | \cdot)$ and $q_{\theta}(\hat{s}_t | \cdot)$, which we will elaborate in the following subsections.

\subsection{Policy Network and Agent Network}
\label{sec4.2}

In Figure \ref{fig:agent}, we show the policy network (top diagram) and the agent network (bottom diagram) in the MMSA method. We adapt the policy and Q-function of SALE \cite{fujimoto2024sale} to model-based MARL settings. Originally, SALE consists of two encoders: a state encoder $f$ and a state-action encoder $g$, where $z^s := f(s)$ is the embedding of $s$, and $z^{sa} := g(z^s, a)$ is the joint state-action embedding \cite{fujimoto2024sale}. 
In our framework, each agent's policy network builds on the SALE state encoder. At time $t$, $s_t$ is passed through the encoder that outputs $z^s_t = f_t(s_t)$. To stabilize downstream learning, we apply AvgL1Norm \cite{fujimoto2024sale}, a normalization function that rescales the input vector and preserves the relative scale of the embedding throughout learning. It can be expressed as
$$\mathrm{AvgL1Norm}\bigl(z^s_t\bigr)
= \frac{z^s_t}{\tfrac{1}{N}\sum_{i=1}^N \lvert z^s_{t,i}\rvert},$$
assuming that $z^s_{t,i}$ is the $i$-th dimension of an $N$-dimensional vector $z^s_t$.
In parallel, the state is mapped through a linear layer to produce $\phi^s_t = \mathrm{Linear}(s_t)$.
The normalized SALE embedding and the learned feature vector 
are then concatenated and fed into the policy head $\pi_t,$
which outputs each agent’s action distribution 
$a^i_t \sim \pi_t\bigl(z^s_t,\;\phi^s_t\bigr)$. 
By decoupling the SALE encoder from the training of the policy, the network benefits from stable state representations \cite{fujimoto2024sale}.

For Q-function estimation, each agent uses its local observation $o^i_t$ and its own action $a^i_t$ at time $t$. First, the SALE state encoder yields 
$z^o_t = f_t\bigl(o^i_t\bigr),$ which is normalized via AvgL1Norm. Then, the state-action encoder computes a joint embedding 
$z^{oa}_t = g_t\bigl(z^{o^i}_t,\, a^i_t\bigr).$ 
In addition, a direct feature mapping of $o^i_t$ and $a^i_t$ is learned via a linear layer
$\phi^{oa}_t = \mathrm{Linear}\bigl(o^i_t,\,a^i_t\bigr)$ and normalized. Lastly, the Q-function for agent $i$ is computed by  
$Q_i\bigl( z^o_t,\; z^{oa}_t,\;\phi^{oa}_t\bigr)$ given the inputs above. 
We follow the training paradigm of the encoders in \cite{fujimoto2024sale}. The encoders are trained concurrently with the agent. However, the gradients from the value function and policy are not propagated to the encoders. We refer to this as a “decoupled” process.
Because the joint state-action embeddings are provided alongside conventional representations, the agent network can exploit rich transition information learned by SALE while maintaining stable training dynamics. 

We use recurrent neural networks (RNNs) to implement the agent network in practice. We denote the hidden outputs of the RNN for agent $i$ and for the whole network as $h^i_t$ and $\mathbf{h}_t$, respectively. We interpret $h^i_t$ as the past knowledge specific to agent $i$ and assume that $\mathbf{h}_t$ collectively encapsulates all the past information of the environment. By this assumption, we can reformulate the approximate posterior $q_{\theta}(\cdot | \hat{s}_{t-1}, \mathbf{a}_{t-1}, \mathbf{o}_{t} )$ as $q_{\theta}(\cdot | \mathbf{h}_t )$. $\mathbf{h}_t$ can be considered as a function of $\hat{s}_{t-1}, \mathbf{a}_{t-1},$ and $ \mathbf{o}_{t}$.
Initially, the posterior latent state is 
$\hat{s}_{t}^{} \sim q_{\theta}(\hat{s}_{t}^{} | \hat{s}_{t-1}, \mathbf{a}_{t-1}, \mathbf{o}_{t}).$
With the reparameterization, it can be transformed into 
$\hat{s}_{t}^{} \sim q_{\theta}(\hat{s}_{t}^{} | \mathbf{h}_{t}).$

\subsection{Integrating World Model and State-Action Learned Embeddings}
\label{sec4.3}

Based on the variational inference analysis and the agent network discussed above, we propose a novel world model that incorporates joint state-action learned embeddings to grasp the underlying dynamics between states and actions. The world model is displayed in Figure \ref{fig:wmodel}. It can be considered as the imagination module in our MMSA framework. This world model significantly enhances sample efficiency by generating imagined roll-outs without the need for interaction with the real environment, thereby saving valuable resources and time. Additionally, the VAEs that estimate prior and posterior distributions of the latent states are integrated into the world model, providing great robustness for probabilistic inference. 
Because we use SALE in our model, the prior distribution learned by the VAE should be written as 
$\hat{s}_t \sim \klprior(\hat{s}_t | {z}_{t-1}^{\hat{s\mathbf{a}}}, z_{t-1}^{\hat{s}}, {\phi}_{t-1}^{\hat{s\mathbf{a}}})$. 

Using Figure \ref{fig:wmodel}, we explain how a complete step of simulation in the world model works. At time step $t$, $\mathbf{h}_{t}$ encapsulates all past information, including the joint action-observation histories $\boldsymbol{\tau}_{t}$, as mentioned in Section \ref{sec4.2}. 
This $\mathbf{h}_{t}$ is passed into both the posterior VAE and the joint agent network (also referred to as the agent networks). The posterior latent state $\hat{s}_{t}^{} \sim q_{\theta}(\hat{s}_{t}^{} | \mathbf{h}_{t})$ is inferred. $\hat{s}_{t}$ is encoded into a state embedding $z^{\hat{s}}_t = f_t(\hat{s}_t)$ and normalized via AvgL1Norm.
On the other hand, the joint agent network is a collection of all agents in the environment, and it implements the policy network that yields the joint policy $\pi := (\pi^1, ..., \pi^N)$ for all agents. Given $\mathbf{h}_{t}$, the joint agent network outputs the joint action 
$\hat{\mathbf{a}}_{t} \sim \pi (\cdot | \boldsymbol{\tau}_{t})$.
$\hat{\mathbf{a}}_{t}$ is passed together with $z^{\hat{s}}_t$ into the state-action encoder to obtain 
$ {z}_{t}^{\hat{s\mathbf{a}}} = g_t ({z}^{\hat{s}}_t, \hat{\mathbf{a}}_t)$.
Meanwhile, a lightweight linear layer also processes $\hat{s}_t$ and $\hat{\mathbf{a}}_{t}$ jointly to produce 
${\phi}_{t}^{\hat{s\mathbf{a}}} = \mathrm{Linear}\bigl(\hat{s}_t, \hat{\mathbf{a}}_{t}\bigr).$
The embeddings $({z}_{t}^{\hat{s\mathbf{a}}}, {z}_{t}^{\hat{s}}, {\phi}_{t}^{\hat{s\mathbf{a}}})$
are passed as inputs to VAE-1, whose decoder predicts the next-step embeddings 
$( {z'}_{t}^{\hat{s\mathbf{a}}}, {z'}_{t}^{\hat{s}}, {\phi'}_{t}^{\hat{s\mathbf{a}}} ).$
Then, we set $\hat{s}_{t+1} = z^{\hat{s}}_{t+1} := {z'}^{\hat{s}}_{t} $, and feed the reconstructed representations $( {z'}_{t}^{\hat{s\mathbf{a}}}, {z'}_{t}^{\hat{s}}, {\phi'}_{t}^{\hat{s\mathbf{a}}} )$ into VAE-2. 
Because these representations embody rich information about the joint state and action at time step $t$, they are sufficient for VAE-2 to derive $\mathbf{h}_{t+1}'$. 
Now, we have both the imagined state $\hat{s}_{t+1}$ and $\mathbf{h}_{t+1}'$ that contains the imagined action-observation histories $\boldsymbol{\tau}_{t+1}'$. Therefore, the one step of imagination is complete, and we are ready to repeat the procedures at time $t+1$.

\subsection{MARL Framework with Joint State-Action Representation Learning}
\label{sec4.4}

Figure \ref{fig:pipeline} presents an overview on the architecture of MMSA. The joint agent network is displayed on the top right, where each agent computes an individual value estimate. Although trained centrally, each agent’s network uses only its own observation and action at execution time. The module on the left shows the world model we described in Section \ref{sec4.3}, with the roll-out horizon set to $j \in \mathbb{Z}$. The roll-out horizon is a tunable hyperparameter. When we perform $j$ simulated roll-outs in the imagination module, we obtain a series of latent states $\{ \hat{s}_{t}, z^{\hat{s}}_{t+1},...,z^{\hat{s}}_{t+j} \}$, or equivalently, $\{ \hat{s}_{t},...,\hat{s}_{t+j} \}$. They are aggregated to form the roll-out state $\hat{s}_{t}^{\text{Rollout}}$. By supplying the aggregated roll-outs to the mixing network, we build a bridge between world model learning and multi-agent value decomposition. 

Combining the world model with the mixing network makes the model applicable in complex multi-agent systems. The MMSA framework operates under the condition of Individual-Global-Max \cite{son_qtran_2019}. Receiving outputs of the joint agent network and merging them monotonically, the mixing network reconciles individual Q-function estimates with the team objective. It models the joint action-value function $Q_{tot}(z^{\mathbf{o}}_{t}, z^{\mathbf{o} \mathbf{a}}_{t}, \phi^{\mathbf{o} \mathbf{a}}_{t})$. This function is consistent with the Q-function under SALE \cite{fujimoto2024sale}, but there is a small difference. Because each of the individual agents in the joint agent network only performs decision-making based on local observation, the individual Q-function should be denoted as $Q_{i}(z^{{o}}_{t}, z^{{o}{a}}_{t}, \phi^{{o} {a}}_{t}),  \forall i \in \mathcal{N}.$ Because MMSA follows the CTDE paradigm, the joint action-value function can be decomposed into individual Q-functions to calculate the expected returns. Therefore, $Q_{tot}$ should also maintain consistency with the $Q_{i}$'s and be a function of joint observation and action, i.e., 
$z^{\mathbf{o}}_{t}, z^{\mathbf{o} \mathbf{a}}_{t}, $ and $\phi^{\mathbf{o} \mathbf{a}}_{t}$.

The mixing network receives three different types of inputs. The first type consists of individual action-value functions from the agent networks, the second is the real global state ${s}_{t}$, and the last is $\hat{s}_{t}^{\text{Rollout}}$. Because $\hat{s}_{t}^{\text{Rollout}}$ incorporates information about potential future observations as well as underlying dynamics between states and actions, it can assist the agents greatly in decision-making.

The overall learning objective consists of four distinct components. First, we minimize the KL divergence between the prior and posterior distributions, $\klprior(\cdot)$ and $q_{\theta}(\cdot)$. Because $\klprior(\cdot)$ is a parameterized surrogate for estimating the true prior, we also need to minimize its divergence from an isotropic Gaussian $p_0(\hat{s}_t) := \mathcal{N}(0, \mathbf{I})$. Combining the two KL terms, we obtain:

\begin{align}
\mathcal{L}_{\text{KL}}(\theta) 
&= \mathcal{D}_{\text{KL}}\left[\klprior \left(\hat{s}_t \mid {z}_{t-1}^{\hat{s\mathbf{a}}}, {z}_{t-1}^{\hat{s}}, {\phi}_{t-1}^{\hat{s\mathbf{a}}} \right) \| p_0(\hat{s}_t)\right] \nonumber \\
& + \mathcal{D}_{\text{KL}}\left[ q_{\theta}(\hat{s}_{t}^{} | \mathbf{h}_{t}) \| \klprior \left(\hat{s}_t \mid {z}_{t-1}^{\hat{s\mathbf{a}}}, {z}_{t-1}^{\hat{s}}, {\phi}_{t-1}^{\hat{s\mathbf{a}}} \right) \right]. \nonumber
\end{align}

We implement the KL balancing technique in the optimization of $\mathcal{L}_{\text{KL}}(\theta)$ because when the learned prior is inaccurate, forcing the posterior to match it aggressively can lead to poor representations \cite{hafner2021mastering}. KL balancing allows the prior to mature quickly while preventing the posterior from being over-regularized. Let $\alpha \in [0, 1]$ be the learning rate. KL balancing can then be defined as:
\begin{align}
\mathcal{D}_{\text{KLB}}\left[q_{\theta}(\cdot) \| \klprior(\cdot)\right]
&= \alpha \mathcal{D}_{\text{KL}} \left[q_{\theta}(\cdot) \; \| \; | \klprior(\cdot) |_{\times} \right] + \nonumber \\
& (1 - \alpha) \mathcal{D}_{\text{KL}}\left[ | q_{\theta}(\cdot) |_{\times} \; \| \; \klprior(\cdot)\right], \nonumber
\end{align}

\noindent where $| \cdot |_{\times}$ denotes the stop-gradient operation.

Second, both the prior and the posterior VAEs are trained to reconstruct the inputs given. We therefore add a loss function measuring how well each VAE recovers the original variables. Using mean‑squared error (MSE), the reconstruction loss is written as:
\begin{align}
\mathcal{L}_{\text{REC}}(\theta) &= \text{MSE}({z}_{t}^{\hat{s\mathbf{a}}}, {z'}_{t}^{\hat{s\mathbf{a}}} ; \theta) + \text{MSE}({z}_{t}^{\hat{s}}, {z'}_{t}^{\hat{s}} ; \theta) + \nonumber \\ 
& \phantom{==} \text{MSE}({\phi}_{t}^{\hat{s\mathbf{a}}}, {\phi'}_{t}^{\hat{s\mathbf{a}}} ; \theta) + \text{MSE}(\mathbf{h}_{t}, \mathbf{h}_{t}' ; \theta). \nonumber
\end{align}


Third, we denote the learnable parameters of the value decomposition framework as $\psi$. Analogous to \cite{rashid2020monotonic}, we define the TD target $y^{tot}$ and TD loss $\mathcal{L}_{\text{TD}}(\psi)$ as follows: 
\begin{align}
& y^{tot} = r_t + \gamma \text{max}_{\mathbf{a}_{t+1}} Q_{tot}(z^{\mathbf{o}}_{t+1}, z^{\mathbf{o} \mathbf{a}}_{t+1}, \phi^{\mathbf{o} \mathbf{a}}_{t+1}, s_{t+1}, \hat{s}_{t+1}^{\text{Rollout}} ; \psi^{-}), \nonumber \\
& \mathcal{L}_{\text{TD}}(\psi) = \left( y^{tot} - 
   Q_{tot}(z^{\mathbf{o}}_{t}, z^{\mathbf{o} \mathbf{a}}_{t}, \phi^{\mathbf{o} \mathbf{a}}_{t}, s_{t}, \hat{s}_{t}^{\text{Rollout}} ; \psi) \right)^2, \nonumber
\end{align}
where $\psi^{-}$ denotes the parameters of the target network.


Fourth, we adapt the MSE loss between the state-action embedding and the embedding of the next state \cite{fujimoto2024sale}. It is applied to train the SALE encoders:
\begin{align}
\mathcal{L}(f, g)
:= \bigl(g_t(f_t(\hat{s}_t), \mathbf{\hat{a}}_{t}) - \lvert f_{t+1}(\hat{s}_{t+1})\rvert_{\times}\bigr)^{2}
= \bigl({z}_{t}^{\hat{s\mathbf{a}}} - \lvert {z}^{\hat{s}}_{t+1}\rvert_{\times}\bigr)^{2}\,. \nonumber
\end{align}

Finally, the overall loss function for MMSA can be written as:
\begin{align}
\mathcal{L}_{\text{total}} = \mathcal{L}_{\text{KL}}(\theta) + \mathcal{L}_{\text{REC}}(\theta) + \mathcal{L}_{\text{TD}}(\psi) + \mathcal{L}(f, g). \nonumber
\end{align}

Through unified optimization of the loss components, MMSA learns to generate accurate latent‐state predictions and improved policies, ultimately maximizing the cumulative return.

\begin{figure*}[ht]
\centering
\begin{subfigure}[t]{.95\linewidth}
  \centering
  \includegraphics[width=\linewidth]{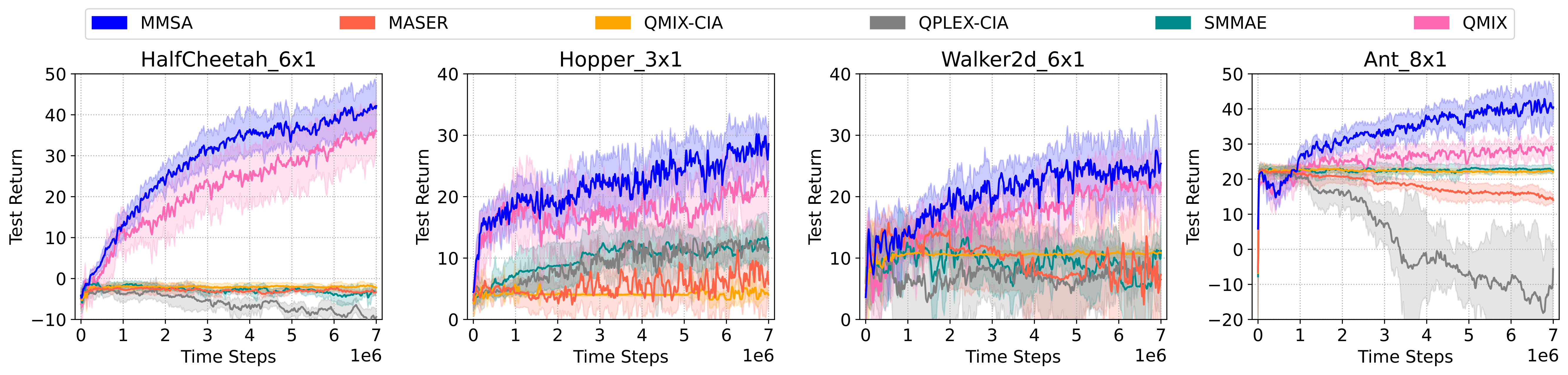}
\end{subfigure}
\begin{subfigure}[t]{.95\linewidth}
  \centering
  \includegraphics[width=\linewidth]{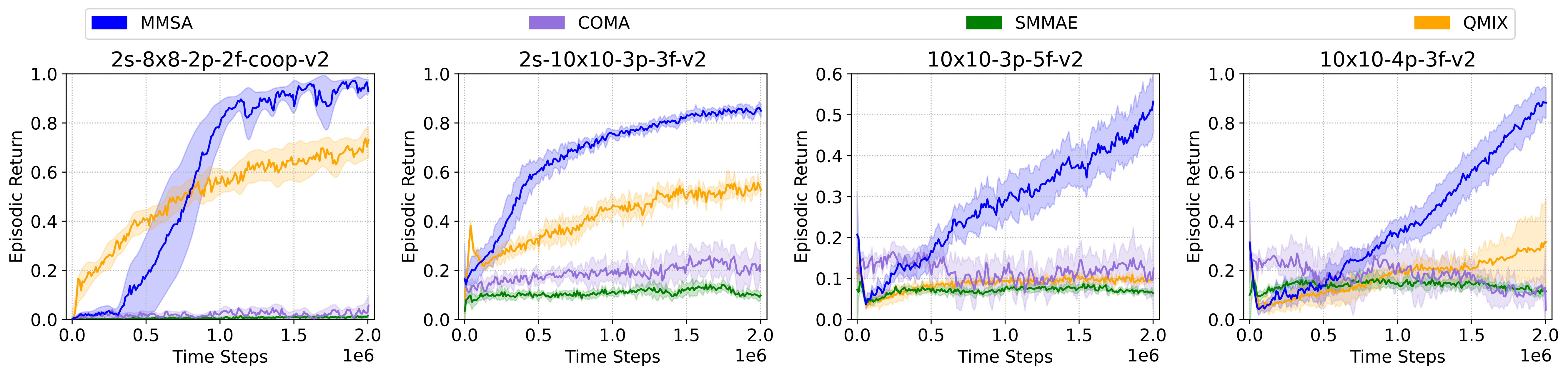}
\end{subfigure}
\caption{Performance of MMSA in Multi-Agent MuJoCo (top row) and in Level-Based Foraging (bottom row). The shaded region captures a $95\%$ confidence interval around the average performance. \textbf{Top: } Comparison of the average episodic return of MMSA with competing MARL algorithms in Multi-Agent MuJoCo tasks. The return is scaled for clear plotting. Experiments are run for 7M time steps. \textbf{Bottom: } The mean episodic return of MMSA compared to other MARL methods in Level-Based Foraging. Each run lasts 2M time steps. In both MARL benchmarks, MMSA excels the competitors in all of the environments. }
\label{fig:mjc_lbf}
\Description{Performance of MMSA in Multi-Agent MuJoCo (top row) and in Level-Based Foraging (bottom row).}
\end{figure*}

\section{Empirical Evaluation}

To evaluate the effectiveness and generalization ability of MMSA, we benchmark our method on various MARL test beds: Multi-Agent MuJoCo (MAMuJoCo) \cite{Peng2020FACMACFM}, Level-Based Foraging (LBF) \cite{Christianos10.5555/3495724.3496622}, and StarCraft Multi-Agent Challenges, including SMAC \cite{samvelyan2019} and SMACv2 \cite{ellis2023smacv2}. MMSA is compared against a broad suite of model-free MARL algorithms (VDN \cite{sunehag_value-decomposition_2017}, COMA \cite{foerster_counterfactual_2018}, QMIX \cite{rashid2020monotonic}, SET-QMIX \cite{li2021permutation}, MASER \cite{pmlr-v162-jeon22a/ALL}, QMIX-CIA \cite{liu2023CIA/ALL}, QPLEX-CIA \cite{liu2023CIA/ALL}, SMMAE \cite{zhang/3545946.3598673/ALL}, HPN-QMIX \cite{jianye2023boosting}, and HPN-VDN \cite{jianye2023boosting}) and model-based MARL methods (MAG \cite{wu2023models}, MAMBA \cite{egorov2022scalable}, and MABL \cite{Venugopal/3635637}). The competing methods range from the classic value-based methods to the newest MARL representatives. All experiments are performed with five different seeds\footnote{The full details of experimental settings can be found in Appendix \ref{appx:expsettings}. Supplementary experiments and results can be found in Appendix \ref{appx:results}.}. 

We then conduct a rigorous design study on the MMSA framework to identify the design elements that most strongly affect the performance. Moreover, we carry out ablation experiments to analyze the impact of different components of MMSA, involving SALE, world model imagination, KL balancing, and the global state. We verify that all components are critical to the competence of MMSA.

\begin{figure*}[htbp]
\centering
\begin{subfigure}{0.7\linewidth}
  \centering
  \includegraphics[width=\linewidth]{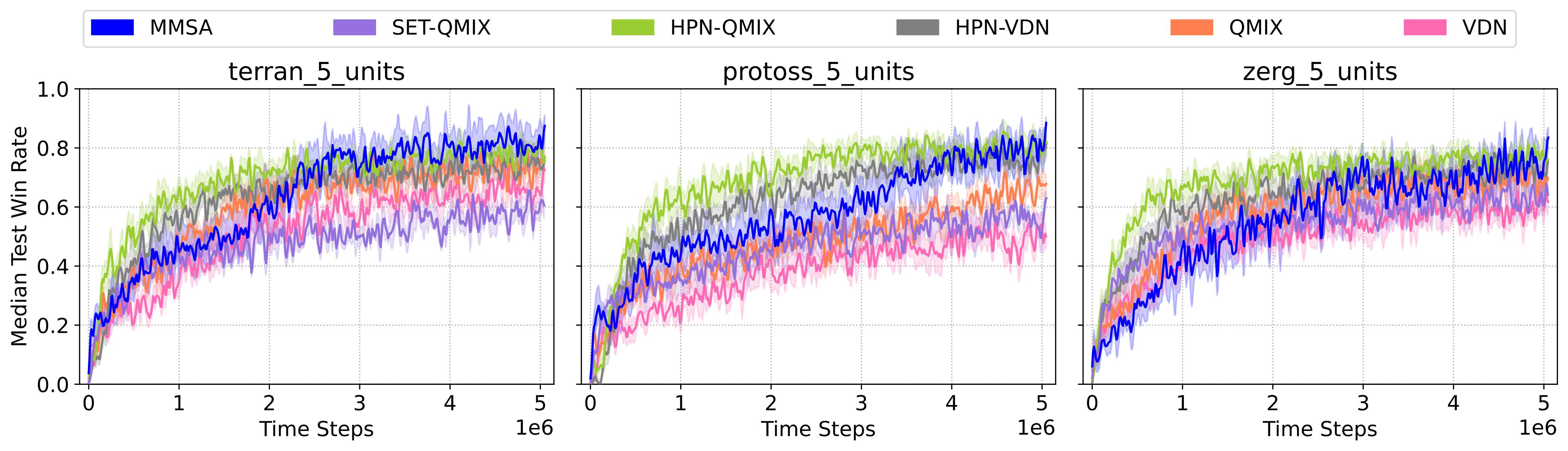}
  \caption{}
  \label{mmsa_smac2}
\end{subfigure}
\begin{subfigure}{0.29\linewidth}
  \centering
  \includegraphics[width=\linewidth]{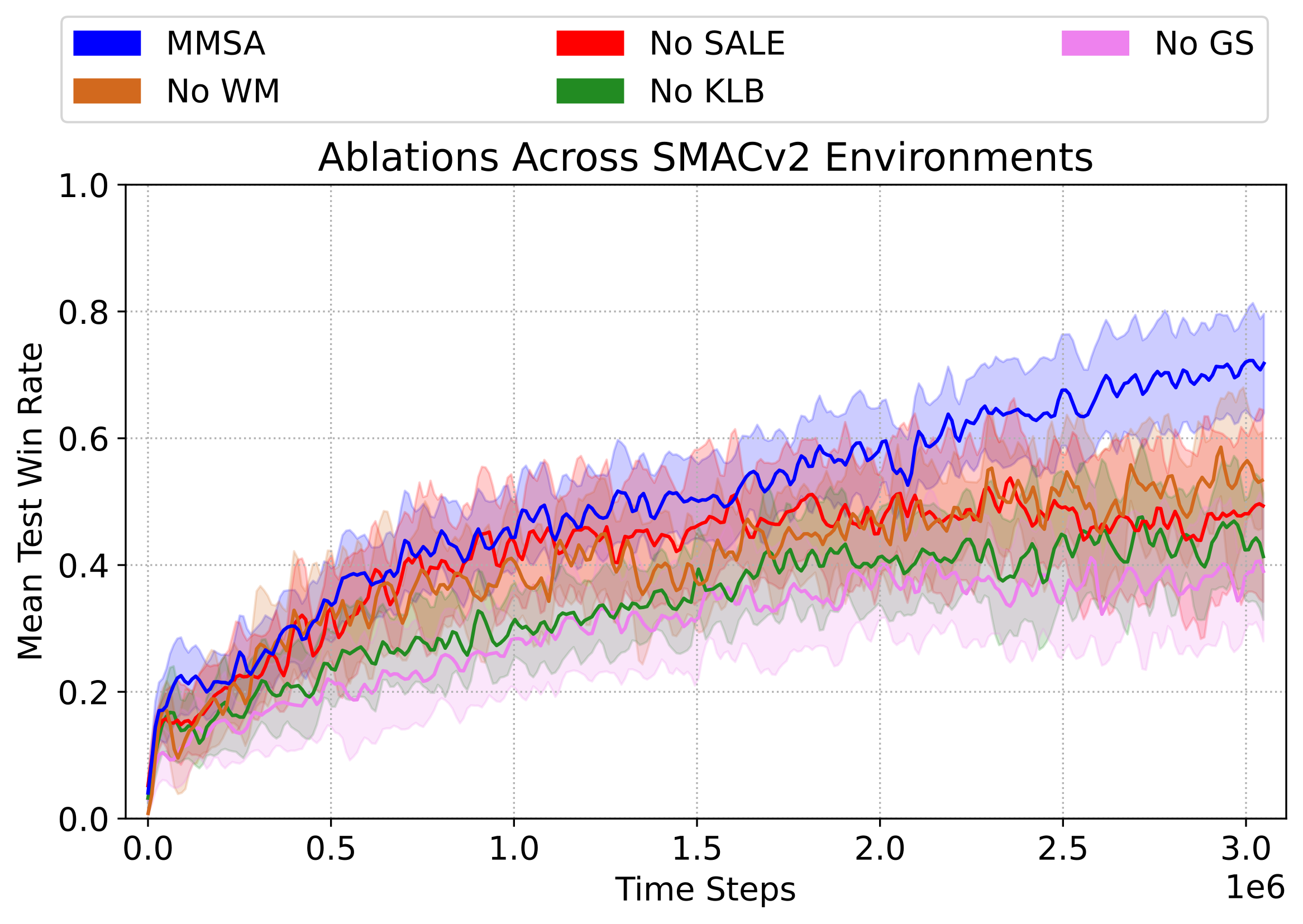}
  \caption{}
  \label{bigabl}
\end{subfigure}
\caption{Performance of MMSA compared with MARL baselines and ablations in SMACv2. (a) Test win rates of MMSA compared with top-performing methods in SMACv2. We plot the median test win rates with the $25\% - 75\%$ percentiles, as in \cite{jianye2023boosting}. Each run lasts 5M time steps. Although MMSA shows a slow start, it gradually outruns the baselines such as VDN and QMIX. It exhibits an overall performance matching that of HPN-QMIX, the best-competing method. (b) Ablations for the MMSA architecture. MMSA is compared against the variants in which the world model, SALE, KL balancing, or global state is removed, respectively (No-WM, No-SALE, No-KLB, or No-GS). Performance is averaged over all SMACv2 challenges. Each run lasts 3M time steps. }
\label{fig:smacv2}
\Description{Evaluation of design choices and ablations of MMSA in SMACv2.}
\end{figure*}

\subsubsection*{Multi-Agent MuJoCo}
To foster the research interest for multi-agent robotic control, \cite{Peng2020FACMACFM} extends the original MuJoCo suite \cite{Todorov6386109} to Multi-Agent MuJoCo (MAMuJoCo) by decomposing a single robot into disjoint sub-graphs. Each of them represents an agent and needs to cooperate to solve continuous control tasks.
As Figure \ref{fig:mjc_lbf} shows, four MAMuJoCo learning tasks with partial observability are completed\footnote{\textit{Ant\_}8\texttimes1 is the Ant partitioned into 8 agents, \textit{Walker2d\_}6\texttimes1 is the Walker partitioned into 6 agents, \textit{HalfCheetah\_}6\texttimes1 is the Half Cheetah partitioned into 6 agents, and \textit{Hopper\_}3\texttimes1 is the Hopper partitioned into 3 agents. In our settings, each agent is constrained to observe only the two nearest joints.}. 
Across all four MAMuJoCo domains, MMSA markedly outperforms the baselines. Several methods, including QPLEX-CIA and MASER, perform poorly and even incur negative returns. This indicates that the robots fail to advance during training. By comparison, MMSA reliably learns forward-moving behaviors for the robot and achieves the highest average return in MAMuJoCo. This superior performance stems from the use of roll-outs generated by the SALE-augmented world model. By simulating and evaluating future joint state-action sequences in latent space, MMSA employs coordinated control strategies that other algorithms, which rely solely on real trajectories, are unable to develop.

\subsubsection*{Level-Based Foraging}
Level-Based Foraging (LBF) \cite{Christianos10.5555/3495724.3496622} is an MARL benchmark that blends cooperation and competition. The general setting of LBF consists of agents and food items. Each of them is assigned an integer level. The agents must coordinate their efforts to collect items whose levels exceed that of any single agent. In our study, we define four distinct LBF environments that vary in the number of agents, the item count, cooperation requirements, partial observability, and the world size. 

Figure \ref{fig:mjc_lbf} displays that MMSA rapidly rises above competing algorithms on every scenario, achieving higher average episodic returns within fewer training episodes. In the task of \textit{10\textup{x}10-3p-5f-v2}, there are three players with five items to collect, more than in any other world. The players need to spend more time gaining rewards. However, MMSA still achieves remarkable progress given the limited time steps. In contrast, QMIX plateaus at lower returns, while other methods, such as MASER and SMMAE, often struggle to coordinate sufficient joint effort and exhibit flatter performance. The results demonstrate that MMSA’s combination of latent imagination, joint representations, and monotonic mixing yields more effective cooperation in the LBF domains.

\begin{figure*}[ht]
\centering
\includegraphics[width=\linewidth]{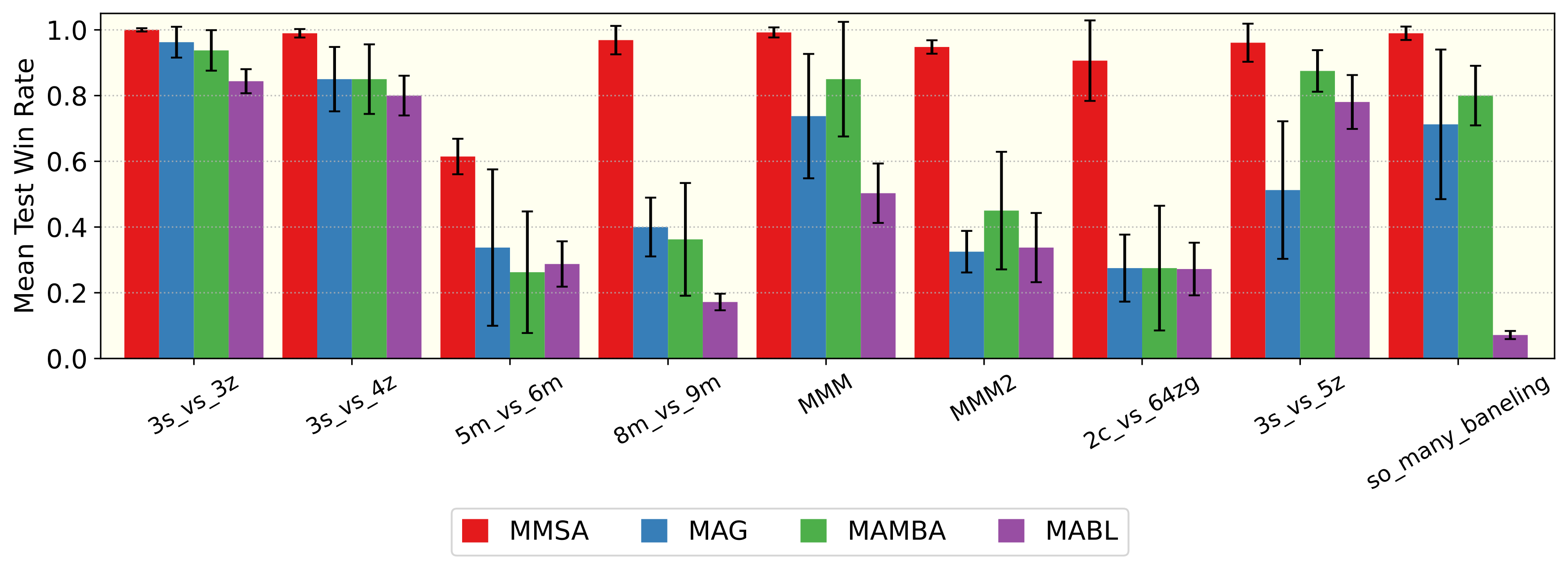}
\caption{Comparing the performance of MMSA with model-based MARL algorithms MAG \cite{wu2023models}, MAMBA \cite{egorov2022scalable}, and MABL \cite{Venugopal/3635637} in StarCraft Multi-Agent Challenges. The error bars represent the 95\% confidence intervals around the mean test win rates.}
\label{fig:mbmarl}
\Description{Comparing the performance of MMSA with model-based MARL algorithms.}
\end{figure*}

\begin{figure}[htbp]
\centering
\includegraphics[width=\columnwidth]{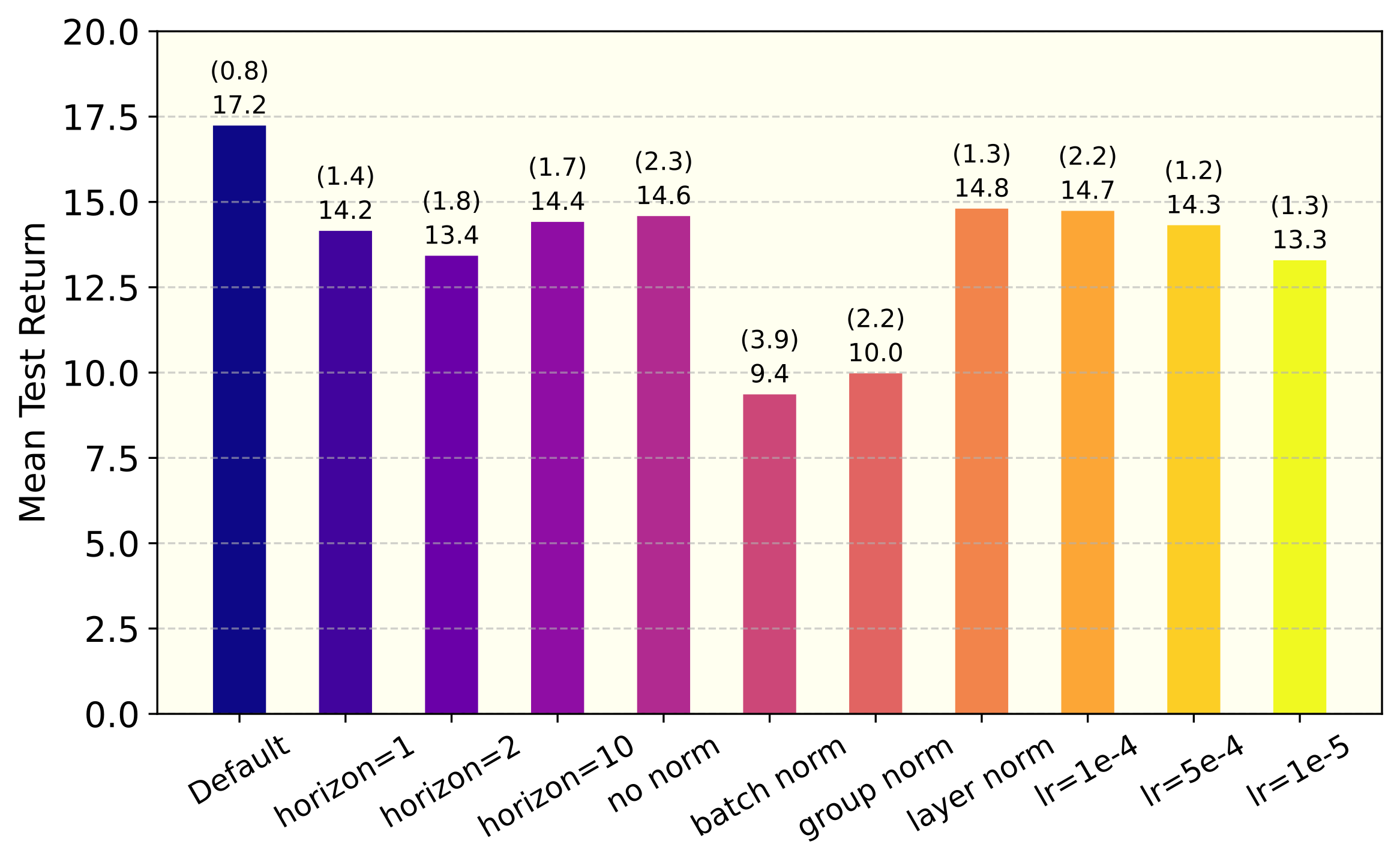}
\caption{We study the design space of the MMSA framework. In the default setting, the roll-out horizon, learning rate, and normalizer are set to 3, 1e-3, and AvgL1Norm, respectively. The mean test return is shown above the bars. Bracketed values stand for the range of the $95\%$ confidence interval around the mean. Experiments are conducted on SMACv2 environments. Each run lasts 3M time steps. The performance is averaged over the SMACv2 scenarios.}
\label{fig:smallabl}
\Description{Evaluation of design choices of MMSA in SMACv2. }
\end{figure}

\subsubsection*{SMAC and SMACv2}

Developed on top of the StarCraft II API, SMAC (SMACv1) comprises a diverse set of battle scenarios corresponding to an extensive range of learning tasks \cite{samvelyan2019}. SMAC targets the problem of micro-management, wherein each agent is responsible for the fine-grained control of an individual unit and selects actions independently.
We conduct experiments on the SMAC benchmark to compare the performance of MMSA with MAMBA (Multi-Agent Model-Based Approach) \cite{egorov2022scalable}, MABL (Multi-Agent Bi-Level world model) \cite{Venugopal/3635637}, and MAG (Models as AGents) \cite{wu2023models}, three model-based MARL algorithms that were recently proposed. Figure \ref{fig:mbmarl} displays the mean and the 95\% confidence interval of the test win rate over five different seeds for each method in SMAC maps. During the experiments, the total number of roll-outs in a single run is the same (2M) for all methods, ensuring that a fair comparison is made. From the results, we can observe that MMSA achieves the highest win rates in every SMAC environment. Furthermore, MMSA exhibits consistency across independent runs, because it yields very small confidence intervals in eight of the nine scenarios.

SMACv2 \cite{ellis2023smacv2} is a stochastic extension of SMAC \cite{samvelyan2019}. Each episode randomizes the allies' and enemies' spawn locations, posing challenges for the allied units to beat the enemies that approach from multiple angles simultaneously. The unit compositions are also randomized, with three unit types per race sampled according to predefined probabilities. 
Figure \ref{mmsa_smac2} shows that MMSA climbs more slowly than the baselines, which could be due to the extra effort required to train the world model with SALE in randomized scenarios. However, MMSA gradually approaches and matches HPN-QMIX across all the SMACv2 maps, surpassing the other methods.
MMSA’s sustained ascent stems from the use of imagined roll-outs. By refining the latent dynamics model and SALE representations, MMSA uncovers more effective coordination strategies over time.

\subsubsection*{Design Studies}
We present detailed studies on three key design elements of MMSA: roll-out horizon, learning rate, and normalization function. Figure \ref{fig:smallabl} implies that our default setting attains the highest average return, demonstrating both strong performance and low variance. 
Reducing the horizon to 1 or 2 steps causes a drop in performance, which indicates that agents may not be able to exploit the full strength of the imagination module with shorter horizons. Increasing the horizon to 10 steps also has a negative impact on the method. Although long roll-outs enable agents to anticipate long-term consequences, the model error can compound over time, leading to unrealistic trajectory predictions as the horizon becomes large. The results confirm that a three-step roll-out should be applied. For normalization, AvgL1Norm proves critical. Omitting normalization or using layer normalization still gives a reasonable return. However, the other normalizers drastically underperform. Lastly, deviating from the default learning rate of 1e-3 yields lower returns. The reason could be that the agents' learning is sensitive to step-size, and smaller learning rates can lead to slower convergence. The default setting proves to be the most effective and reliable design choice.

\begin{table}[h]
  \centering
  \caption{An overview of the performance of MMSA at the end of training compared to competing methods across different MARL benchmarks. For SMAC (SMACv1) and SMACv2, we use the mean test win rate averaged over the environments, instead of the median win rate. For MAMuJoCo and LBF, the episodic returns are averaged over all tasks within the benchmarks. MMSA leads the group among MAMuJoCo, LBF, and SMAC. On SMACv2, MMSA ties with HPN-QMIX and outperforms the other MARL approaches. Refer to Appendix \ref{appx:results} for supplementary results.}
  \vspace{0.3em}
  \begin{subtable}{\columnwidth}
    \centering
    \begin{tabular*}{.9\columnwidth}{cccc}
    \hline
        Environments & \footnotesize{MMSA} & \footnotesize{MASER} & \footnotesize{SMMAE} 
        \\ \hline 
        MAMuJoCo Tasks & \textbf{36.94} & 5.94 & 11.01  
        \\ \hline
        Environments & \footnotesize{QMIX-CIA} & \footnotesize{QPLEX-CIA} & \footnotesize{QMIX} 
        \\ \hline 
        MAMuJoCo Tasks & 8.71 & 1.85 & 28.69 
        \\ \hline
    \end{tabular*}
    \label{tab:sub1}
  \end{subtable}
  \vspace{0.2em}

  \begin{subtable}{\columnwidth}
    \centering
    \begin{tabular*}{.9\columnwidth}{ccccc}
    \hline
        Environments & \footnotesize{MMSA} &  \footnotesize{SMMAE} & \footnotesize{QMIX} & \footnotesize{COMA} \\ \hline 
        LBF Environments & \textbf{0.80} & 0.08 & 0.42 & 0.11 \\
        \hline
    \end{tabular*}
    \label{tab:sub2}
  \end{subtable}
  \vspace{0.2em}

  \begin{subtable}{\columnwidth}
    \centering
    \begin{tabular*}{.9\columnwidth}{ccccc}
    \hline
        Environments & \footnotesize{MMSA} &  \footnotesize{MAG} & \footnotesize{MAMBA} & \footnotesize{MABL} \\ \hline 
        SMACv1 Challenges & \textbf{0.93} & 0.57 & 0.63 & 0.45 \\
        \hline
    \end{tabular*}
    \label{tab:sub3}
  \end{subtable}
  \vspace{0.2em}

  \begin{subtable}{\columnwidth}
    \centering
    \begin{tabular*}{.9\columnwidth}{cccc}
    \hline
        Environments & \footnotesize{MMSA} & \footnotesize{SET-QMIX} & \footnotesize{HPN-QMIX} 
        \\ \hline 
        SMACv2 Challenges & \textbf{0.81} & 0.64 & \textbf{0.81}  
        \\ \hline
        Environments & \footnotesize{HPN-VDN} & \footnotesize{QMIX} & \footnotesize{VDN} 
        \\ \hline 
        SMACv2 Challenges & 0.78 & 0.72 & 0.58 
        \\ \hline
    \end{tabular*}
    \label{tab:sub4}
  \end{subtable}

  \label{tab:four_metrics}
\end{table}

\subsubsection*{Ablation Studies}
Moreover, to study the contributions of each architectural component, we compare the complete MMSA framework against four ablated variants: without the world model (No-WM), without using SALE (No-SALE), without KL balancing (No-KLB), and without global state in the mixing network (No-GS). The results are displayed in Figure \ref{bigabl}. Firstly, removing the learned world model caps the win rate around 0.5, which indicates that the world model’s roll-out predictions are vital for pushing the agents beyond baseline behaviors. Secondly, omitting the SALE mechanism yields a lower asymptotic performance than No-WM, which highlights the importance of modeling the underlying structure of the environment and capturing the interactions between states and actions. Thirdly, the learning curve of No-KLB reveals the significance of KL balancing in preventing posterior collapse and maintaining robust representation learning. Finally, the evident performance drop of No-GS implies that access to the centralized state during training is critical for effective coordination in SMACv2 scenarios. By combining these key components, MMSA achieves the greatest efficiency in early learning and the highest asymptotic win rate.

\section{Conclusion}

This paper presents MMSA, a model-based MARL method that fuses a value factorization framework with joint state-action representation learning, amortized variational inference, and an imagination module. MMSA is able to produce faithful latent roll‑outs, preserve well‑scaled embeddings, and learn decentralized policies from real and imagined experience. Experiments on various MARL benchmarks demonstrate the outstanding performance and generalizability of our approach. Design studies justify the design choices for the MMSA method. Ablation studies confirm the positive impact and indispensability of different MMSA components.

A promising avenue for future research is the systematic mitigation of model error, which is a long-standing problem for model-based MARL. The discrepancy between imagined roll‑outs and real‑world dynamics may accumulate and ultimately misguide cooperative policies. Extending MMSA with an ensemble of models could mitigate the impact of the errors, as model ensembles have proven to be effective in reducing model uncertainties. Alternatively, incorporating uncertainty estimation techniques and applying regularization schemes could prevent the model from overfitting and promote generalization to unseen states.

\bibliographystyle{ACM-Reference-Format} 
\bibliography{references}

\newpage
\onecolumn
\appendix
\newpage
\section{Supplementary Results}
\label{appx:results}

\subsection{Additional Results in SMACv2 Scenarios}

\begin{figure}[htbp]
\centering
\includegraphics[width=\linewidth]{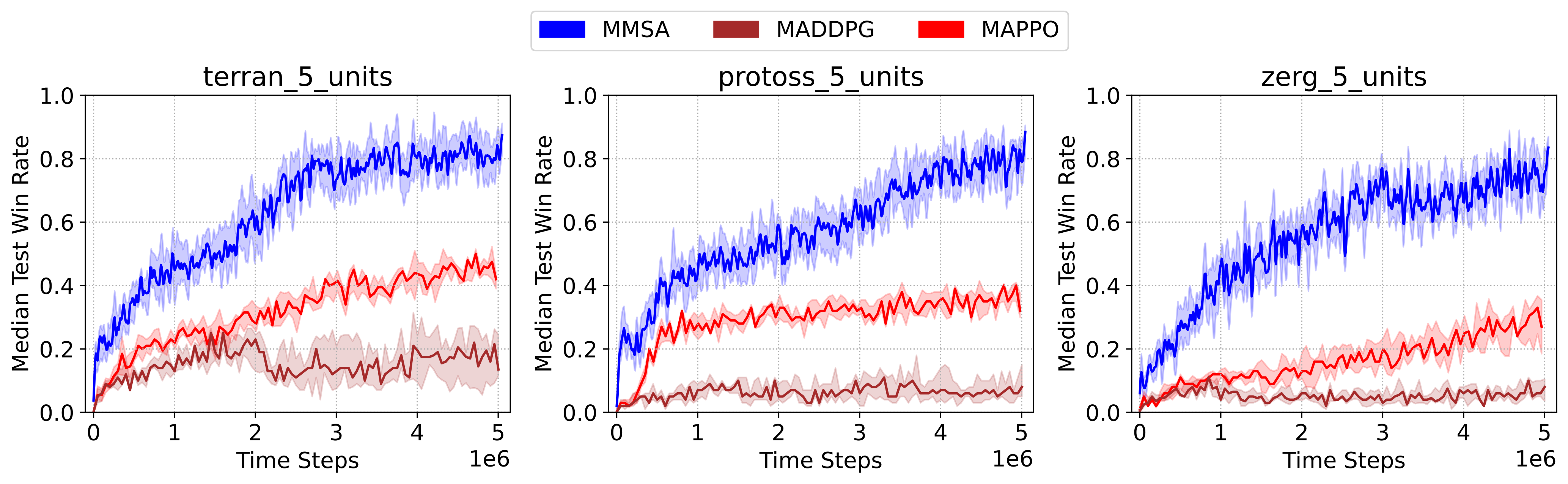}
\caption{Comparison of the median test win rate of MMSA against additional MARL methods, including MADDPG \cite{10.5555/Lowemaddpg} and MAPPO \cite{10.5555/yumappo}, in SMACv2 battle scenarios. The shaded region represents the 25\% - 75 \% percentiles. Each run lasts 5M time steps. }
\Description{MMSA median win rates extra results}
\end{figure}

\subsection{Additional Results in Multi-Agent MuJoCo Environments}

\begin{figure}[htbp]
\centering
\includegraphics[width=\linewidth]{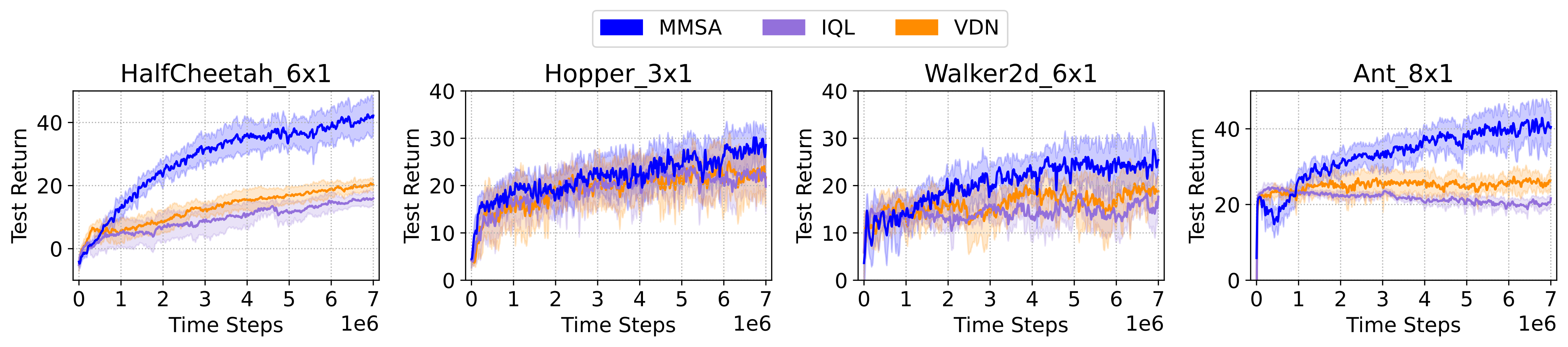}
\caption{Episodic return of MMSA against additional MARL algorithms, including VDN \cite{sunehag_value-decomposition_2017} and IQL \cite{tampuu_multiagent_2017}, in Multi-Agent MuJoCo environments. The shaded area captures a 95\% confidence interval around the mean performance. For clear plotting, the values of the returns are scaled for all methods, and the details can be found in Appendix \ref{appx:expsettings}. }
\Description{Episodic return of MMSA (1)}
\end{figure}

\subsection{Additional Results in Level-Based Foraging Tasks}

\begin{figure}[H]
\centering
\includegraphics[width=\linewidth]{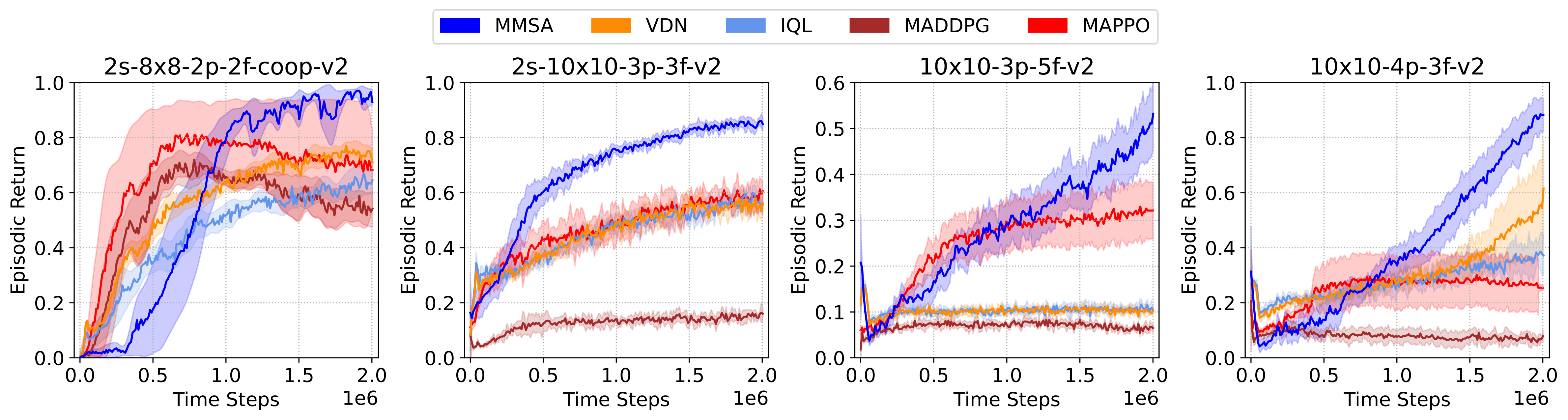}
\caption{Episode return of MMSA compared with the MARL baselines, including VDN, IQL, MADDPG \cite{10.5555/Lowemaddpg}, and MAPPO \cite{10.5555/yumappo}, in Level-Based Foraging environments. We select four representative tasks in which the levels of partial observability and cooperation vary, and in which the numbers of players and food items differ. The shaded area captures a 95\% confidence interval around the mean performance.}
\Description{Episodic return of MMSA (2)}
\end{figure}

\subsection{Overview of the Final Performance Comparison Between MMSA and the Baseline Algorithms}

\begin{table}[H]
    \centering
    \caption{An overview of the performance of MMSA compared to SET-QMIX, HPN-QMIX, HPN-VDN, QMIX, VDN, MADDPG, MAPPO, MAMBA, and MAG across the SMACv2 scenarios. The performance is measured as the mean of the test win rates.}
    \vspace{1em}
    \begin{tabular}{ccccccccccc}
    \hline
        \footnotesize{SMACv2 Maps$\backslash$Methods} & {MMSA} & {SET-QMIX} & {HPN-QMIX} & {HPN-VDN} & {QMIX} & {VDN} & {MADDPG} & {MAPPO} & {MAMBA} & {MAG} \\ \hline 
        \textit{terran\_5\_units} & 0.83 & 0.64 & 0.82 & 0.70 & 0.75 & 0.67 & 0.20 & 0.45 & 0.54 & 0.59 \\ 
        \textit{protoss\_5\_units} & 0.81 & 0.63 & 0.84 & 0.84 & 0.69 & 0.46 & 0.10 & 0.38 & 0.54 & 0.58 \\ 
        \textit{zerg\_5\_units}& 0.81 & 0.64 & 0.78 & 0.79 & 0.73 & 0.62 & 0.09 & 0.29 & 0.32 & 0.34 \\
        \hline
    \end{tabular}
\label{tbl:supresults1}
\end{table}

\begin{table}[H]
     \caption{An overview on the final average episodic return of MMSA compared with other MARL baselines in Multi-Agent MuJoCo learning tasks. Among all the competing baselines, QMIX and VDN acquire the highest episodic return at the end of the testing episodes. MMSA surpasses QMIX by 29\% and VDN by 65\% across different MAMuJoCo tasks on average.}
     \vspace{1em}
    \begin{tabular}{ccccccccc}
    \hline
       {MAMuJoCo Tasks$\backslash$Methods} & {MMSA} & {MASER} & {IQL} & {QMIX-CIA} & {QPLEX-CIA} & {SMMAE} & {QMIX} & {VDN} \\ \hline
       \textit{HalfCheetah}\_6\texttimes1 & 43.10 & -3.14 & 16.03 & -2.18 & -9.26 & -3.28 & 35.83 & 20.13 \\ 
        \textit{Hopper}\_3\texttimes1 & 31.02 & 5.86 & 17.09 & 3.94 & 12.18 & 12.52 & 22.09 & 24.10 \\ 
        \textit{Ant}\_8\texttimes1 & 45.06 & 13.65 & 22.50 & 22.23 & -2.09 & 22.62 & 33.50 & 26.73 \\ 
        \textit{Walker2d}\_6\texttimes1 & 28.56 & 7.40 & 18.61 & 10.86 & 6.56 & 12.16 & 23.33 & 18.72 \\ 
        \hline
    \end{tabular}
\label{tbl:supresults2}
\end{table}

\begin{table}[H]
    \centering
    \caption{An overview on the final episode return of MMSA compared with the other MARL approaches in Level-Based Foraging. Our method makes substantial performance gains across distinct LBF tasks compared to QMIX, VDN, MAPPO, and IQL. }
    \vspace{1em}
    \begin{tabular}{ccccccccccc}
    \hline
        \vspace{0.2em}
        LBF Environments$\backslash$Methods & MMSA & COMA & SMMAE & QMIX & VDN & IQL & MADDPG & MAPPO & IPPO & MAA2C \\ \hline 
        \textit{10x10-3p-5f-v2} & 0.53 & 0.12 & 0.07 & 0.11 & 0.10 & 0.11 & 0.07 & 0.33 & 0.50 & 0.45 \\ 
        \textit{10x10-4p-3f-v2} & 0.88 & 0.04 & 0.12 & 0.31 & 0.57 & 0.37 & 0.08 & 0.27 & 0.58 & 0.60\\ 
        \textit{2s-8x8-2p-2f-coop-v2} & 0.93 & 0.06 & 0.02 & 0.73 & 0.71 & 0.65 & 0.54 & 0.69 & 0.76 & 0.77\\ 
        \textit{2s-10x10-3p-3f-v2} & 0.86 & 0.20 & 0.10 & 0.53 & 0.56 & 0.56 & 0.16 & 0.61 & 0.57 & 0.53\\
        \hline
    \end{tabular}
\label{tbl:supresults3}
\end{table}

\begin{table}[ht]
  \centering
  \caption{An overview of the average run time (in hours) of MMSA compared with baselines on MAMuJoCo, LBF, SMAC, and SMACv2. All algorithms are run for the same number of time steps on each benchmark.}
  \vspace{0.3em}

  \begin{subtable}{\linewidth}
    \centering
    \begin{tabular}{cccccccc}
    \hline
        Environment & {MMSA} & {SMMAE} & {MASER} & {QMIX-CIA} & {QPLEX-CIA} & {QMIX} & {VDN}
        \\ \hline
        MAMuJoCo & 33.3 & 33.0 & 41.4 & 20.8 & 18.7 & 16.5 & 16.7
        \\ \hline
    \end{tabular}
    \label{rt1}
  \end{subtable}
  \vspace{1em}

  \begin{subtable}{\linewidth}
    \centering
    \begin{tabular}{ccccccc}
    \hline
        Environment & {MMSA} & {COMA} & {SMMAE} & {MADDPG} & {MAPPO} & {QMIX} \\ \hline
        LBF & 3.7 & 0.7 & 4.3 & 1.8 & 0.9 & 2.1 \\
        \hline
    \end{tabular}
    \label{rt2}
  \end{subtable}
  \vspace{1em}

  \begin{subtable}{\linewidth}
    \centering
    \begin{tabular}{ccccc}
    \hline
        Environment & {MMSA} &  {MAG} & {MAMBA} & {MABL} \\ \hline
        SMACv1 & 14.6 & 63.9 & 68.7 & 37.5 \\
        \hline
    \end{tabular}
    \label{rt3}
  \end{subtable}
  \vspace{1em}

  \begin{subtable}{\linewidth}
    \centering
    \begin{tabular}{cccccccc}
    \hline
        Environment & {MMSA} & {SET-QMIX} & {HPN-QMIX} & {MADDPG} & {MAPPO} & {HPN-VDN} & {QMIX}
        \\ \hline
        SMACv2 & 19.7 & 10.9 & 19.9 & 9.9 & 4.4 & 19.3 & 16.8
        \\ \hline
    \end{tabular}
    \label{rt4}
  \end{subtable}

  \label{tab:runtimes}
\end{table}

\subsection{Comparing Losses of the Alternate Design Choices of MMSA}

\begin{figure}[h]
\centering
\includegraphics[width=\linewidth]{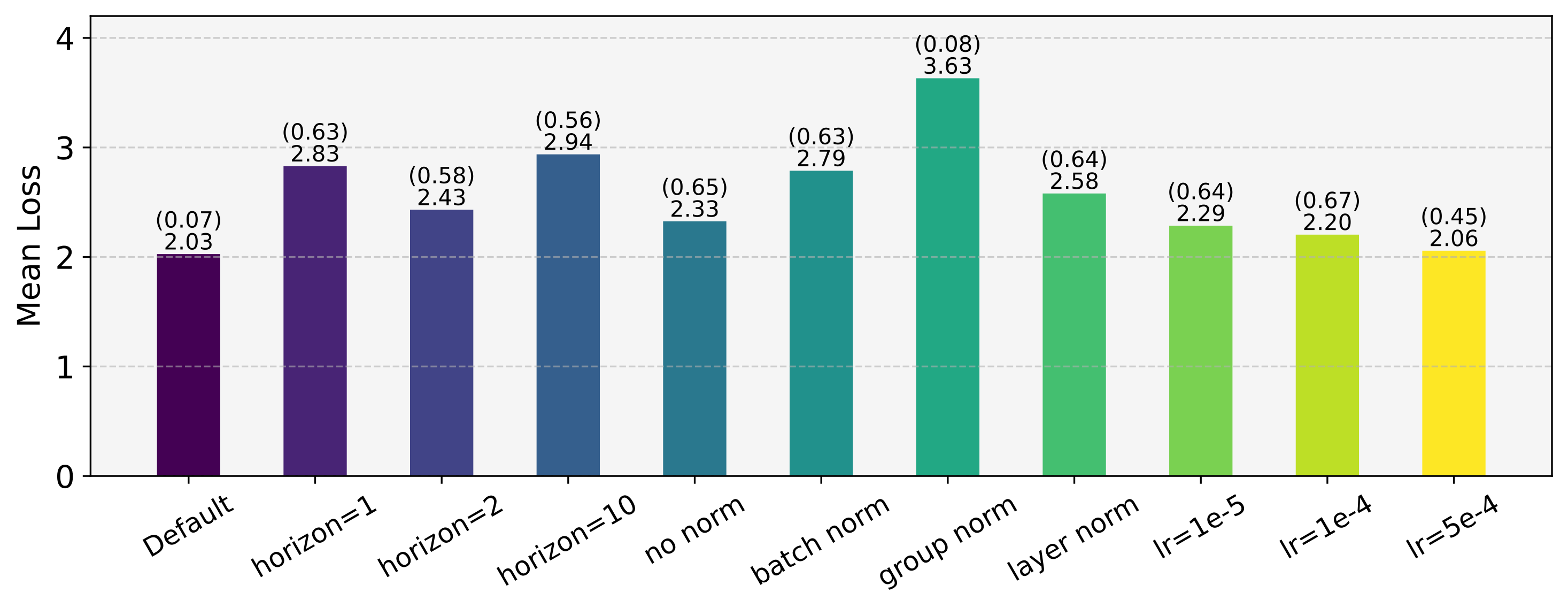}
\caption{We conduct detailed design studies for the MMSA framework. In the default setting of MMSA, the roll-out horizon is set to 3, the learning rate is set to 1e-3, and the normalization function is set to AvgL1Norm. This figure shows the mean loss from using alternate design choices for MMSA. The results are collected from experiments across the SMACv2 environments. Each run lasts 3M time steps. The mean loss values are displayed above the bars. The numbers in brackets represent the 95\% confidence interval around the mean loss. }
\Description{Loss of MMSA and design space}
\label{fig:supploss}
\end{figure}

Figure \ref{fig:supploss} displays the mean loss when we use different design choices for MMSA. It demonstrates that our default setting of MMSA achieves the smallest loss over the design space. The loss can potentially be further optimized if we extend MMSA with an ensemble of models or incorporate uncertainty estimation techniques into the framework, which are the future research directions mentioned before.

\subsection{Numerical Comparison Between MMSA and Model-Based MARL Baselines}

\begin{figure}[htbp]
\centering
\includegraphics[width=\linewidth]{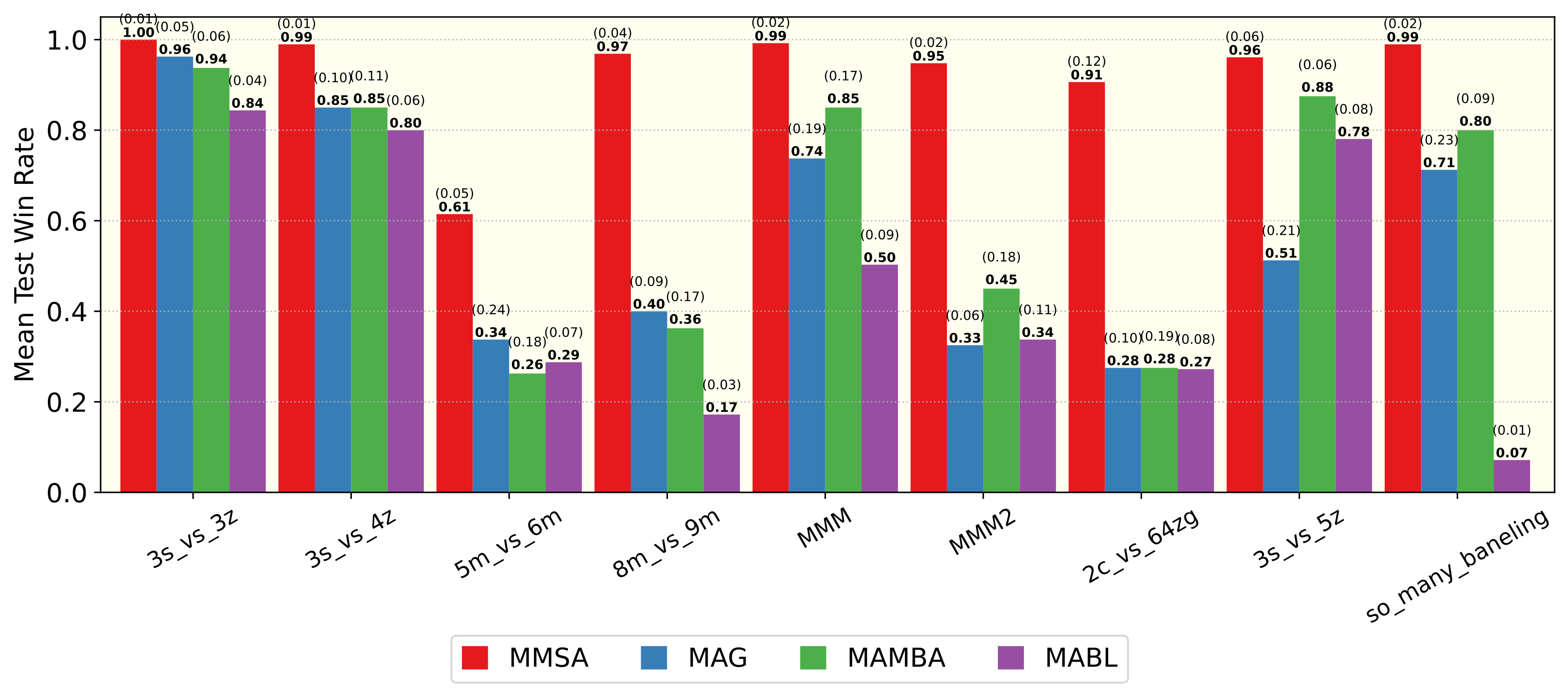}
\caption{Comparing the performance of MMSA with model-based MARL baselines in StarCraft Multi-Agent Challenges. The number in bold represents the mean of test win rates. The number in brackets stands for the 95\% confidence interval around the mean performance.}
\Description{ MMSA with model-based MARL baselines}
\label{fig:mbmarlsupp}
\end{figure}

We conduct experiments on the StarCraft Multi-Agent Challenge (SMACv1) benchmark to compare the performance
of MMSA with three model-based MARL algorithms, including MABL (Multi-Agent Bi-Level world model) \cite{Venugopal/3635637}, MAMBA (Multi-Agent Model-Based Approach) \cite{egorov2022scalable}, and MAG (Models as AGents) \cite{wu2023models}. Figure \ref{fig:mbmarlsupp} displays the mean and the 95\% confidence interval of the test win rate over five different seeds for each method in SMACv1 maps. During the experiments, the total number of roll-outs in a single run is the same (2M) for all three methods, ensuring that a fair comparison is made. 

\newpage
\section{Overview of the Benchmark Environments}
\label{appx:envs}

\subsection{StarCraft Multi-Agent Challenges (SMACv1 and SMACv2)}

Our algorithm is tested in the StarCraft Multi-Agent Challenges (SMAC, also known as SMACv1) \cite{samvelyan2019} and SMACv2 \cite{ellis2023smacv2}, an improved benchmark built on SMACv1. Designed using the popular RTS game StarCraft II, SMACv1 and SMACv2 foster the development of MARL methods in complex and real-time settings. They encourage the performance analysis of diverse MARL algorithms on standardized benchmarks.

The micro-management tasks of SMACv1 vary in terms of the number and types of units, creating a diverse set of challenges that require effective cooperation and strategies. For example, although \textit{bane\_vs\_bane} is symmetric, it has 24 units in one team, and 4 of them are Banelings. They are suicide bomber units that explode when being killed, which increases the difficulty of battles. 

The 7 allied Zealots in \textit{so\_many\_baneling} have to survive the attacks by 32 enemy Banelings that are strong against Zealots. They must design a plan for positioning, spreading far from each other on the terrain so that the Banelings' suicidal attacks inflict minimal damage on them. In \textit{3s\_vs\_4z} and \textit{3s\_vs\_5z}, the allied Stalkers need a specific strategy called kiting in order to vanquish the enemy Zealots, which causes delays to the reward. 

SMACv2 \cite{ellis2023smacv2} introduces procedural generation of map layouts, randomized unit compositions, and strict field-of-view constraints, thereby restoring the need for adaptive, closed-loop coordination and robust generalization. The core of SMACv2 lies in several orthogonal modifications to the original SMAC environments:

(1) Randomized Start Positions. Two distinct spawn patterns prevent memorization of fixed openings and force agents to adapt tactics per episode. The first is \textit{surrounded}, where allied units appear centrally and are encircled on multiple flanks by enemy forces. The second is \textit{reflect\_position}, where allied units are spawned uniformly and their positions are mirrored about the map center for the placement of enemies.

(2) Probabilistic Unit Types. Rather than fixed homogeneous armies, SMACv2 samples unit types from pre-specified race-dependent distributions (three types per Protoss, Terran, and Zerg race), ensuring varied team compositions that exercise both strategic planning and fine-grained micro-skills.

(3) Conical Partial Observability. Agents’ sight and attack ranges can be adjusted to a forward-facing cone, aligning more closely with true StarCraft II mechanics and compelling active information gathering.

By procedurally generating map features and observation limitations, SMACv2 draws evaluation scenarios from the same underlying distribution but with unpredictable layouts, breaking the overfit-friendly regime of SMAC.

In our experiments, the initial learning phase of MMSA is slower compared to the model-free methods HPN-QMIX and HPN-VDN \cite{jianye2023boosting}. This can be a consequence of our model-based method interacting with stochastic SMACv2 scenarios. Since SMACv2 has randomized spawn points and unit types, it introduces significant uncertainty into the transition dynamics. MMSA learns an internal world model to simulate trajectories. In highly stochastic environments like SMACv2, the model requires a larger number of samples to accurately capture the distribution of environment dynamics before it can generate reliable imaginary rollouts. In contrast, MMSA excels at sample efficiency and asymptotic performance in environments with low or no stochasticity.

\begin{figure}[htbp]
\centering
\includegraphics[width=\linewidth]{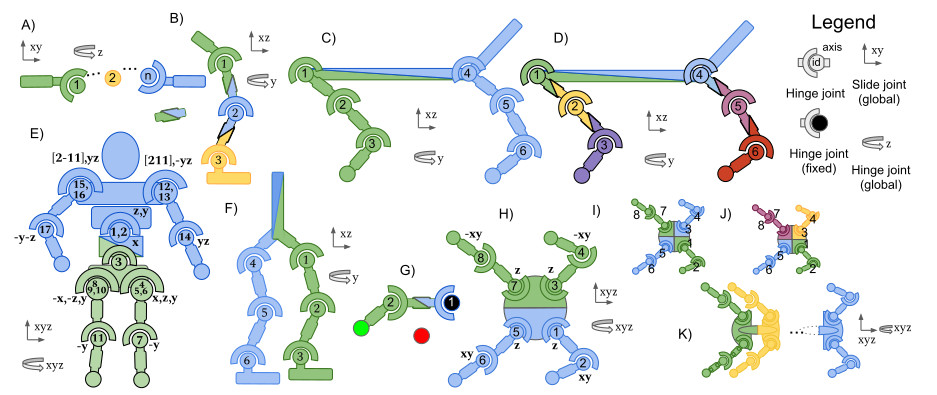}
\caption{Different configurations of robots in Multi-Agent MuJoCo. A) Many-Agent Swimmer; B) Three-Agent Hopper; C) Two-Agent Half Cheetah; D) Six-Agent Half Cheetah; E) Two-Agent Humanoid; F) Two-Agent Walker2d; G) Two-Agent Reacher; H) Two-Agent Ant; I) Diagonal Two-Agent Ant; J) Four-Agent Ant; K) Many-Agent Ant. The figure is extracted from (\cite{Peng2020FACMACFM}, p. 12214).}
\Description{Different configurations of robots in Multi-Agent MuJoCo.}
\label{fig:suppmjc1}
\end{figure}

\subsection{Multi-Agent MuJoCo}

The Multi-Agent MuJoCo (MAMuJoCo) environment, developed by \cite{Peng2020FACMACFM}, is a benchmark for multi-agent robotic control and learning. It is based on OpenAI MuJoCo environments, which have been widely applied in the research area of single-agent RL. 

As demonstrated by Figure \ref{fig:suppmjc1}, a single robot in the MAMuJoCo environment can be partitioned into disjoint sub-graphs, each of which represents an agent. The agent may control one or more joints, depending on the specific agent partitioning. Each of the joints corresponds to a controllable motor. All agents need to collaborate on solving distinct tasks. In this environment, agents can be configured with different levels of observational capabilities. For instance, an agent can be set to observe only the state of its own joints and body, or it can be set to observe its immediate neighbor’s joints and bodies.

The MAMuJoCo configurations used in this paper include:
\begin{itemize}
    \item The Three-Agent partitioning of Hopper denoted as \textit{Hopper\_}3\texttimes1 (configuration B in Figure \ref{fig:suppmjc1}),
    \item The Six-Agent partitioning of Half Cheetah denoted as \textit{HalfCheetah\_}6\texttimes1 (configuration D in Figure \ref{fig:suppmjc1}),
    \item The Six-Agent partitioning of Walker2d denoted as \textit{Walker2d\_}6\texttimes1 (every agent controls a joint of Walker2d),
    \item The Eight-Agent partitioning of Ant denoted as \textit{Ant\_}8\texttimes1 (every agent controls a joint of Ant).
\end{itemize}

In our implementation, we discretize the action space of MAMuJoCo to accommodate agents designed for discrete actions. Specifically, the action space is discretized into $N$ equally spaced atomic actions, where $N$ is an adjustable number. A robot can be partitioned into different sub-graphs, and its joints are connected by adjacent edges. Agents can be configured with different observation constraints. All features in the states are normalized.

\subsection{Level-Based Foraging}

The Level-Based Foraging (LBF) environment \cite{Christianos10.5555/3495724.3496622, Papoudakis2020BenchmarkingMD} is a collection of mixed cooperative-competitive games in the field of multi-agent reinforcement learning. In this environment, agents and food items are randomly placed in a grid world. Each of them is assigned a level, and the level may vary. The task for each agent is to navigate the grid world map and collect the food items. In order to load the items, agents have to choose a certain action next to the item. However, such a collection is only successful if the sum of the levels of agents involved in loading food is equal to or greater than the food item level. Agents receive a reward equal to the level of the collected food divided by their contribution (their levels). LBF provides a test bed for studying both cooperation and competition among agents in MARL. In our experiments, four distinct LBF tasks are defined, with variable environment configurations. 

To customize the environment configuration, \cite{Christianos10.5555/3495724.3496622} provide a template: "Foraging\{obs\}-\{x\_size\}x\{y\_size\}-\{n\_agents\}p-\{food\}f\{force\_c\}-v2". The options in the template are as follows:
\begin{itemize}
    \item \{obs\}: This option introduces partial observability into the environment. For example, if it is set to "-2s", each agent will only have a visibility radius of 2. Otherwise, it can be left blank.
    \item \{x\_size\} and \{y\_size\}: These fields determine the size of the grid world. For example, an $8\times8$ grid world is expressed as "8x8" in the template.
    \item \{n\_agents\}: This option sets the number of agents (players) for the task.
    \item \{food\}: This field sets the number of food items that will be randomly placed in the grid world.
    \item \{force\_c\}: This field indicates whether the task will be fully cooperative or not. If it is set to "-coop", then the environment will only contain food items that require the cooperation of all agents to be collected successfully.
\end{itemize}

For example, we have created an LBF environment named "\textit{2s-8\textup{x}8-2p-2f-coop-v2}" for the experiments. This configuration indicates that there are two players and two food items randomly scattered in an $8\times8$ grid world. Every player has a visibility radius of two. They have to fully collaborate on collecting the food.

\section{Experimental Settings}
\label{appx:expsettings}

In Table \ref{tab:mmsaparams} and Table \ref{tab:smacparams}, we show the hyperparameter settings for MMSA and the environmental set-up for SMACv2, respectively. The probability distributions of randomized units by race are displayed in Table \ref{tab:smacprobs}. The settings for Multi-Agent MuJoCo and for Level-Based Foraging are exhibited in Table \ref{tab:mujocoparams} and Table \ref{tab:lbfparams}. To make clear graphs for Multi-Agent MuJoCo, we scale the returns for the four tasks in MAMuJoCo. We have specified this in Table \ref{tab:mujocoparams}.
The hyperparameters of the other MARL baselines are consistent with their official implementations. Our implementation for MMSA is based on PyTorch.

\begin{table}[htbp]
\centering
\caption{The set-up of hyperparameters for our MMSA method.}
\vspace{1em}
\begin{tabular}{|l | l|}
\hline
\textbf{Name}                                        & \textbf{Value} \\ \hline

Action selector                    & Epsilon-greedy \\
Agent architecture                  & RNN \\
Batch size                          & 32  \\
Buffer size                         & 5000 \\
Number of environments to run in parallel       & 1  \\
Learning rate                       & $1 \times 10^{-3}$ \\
Discount hyperparameter $\gamma$    & 0.99 \\
Optimizer                           & RMSProp       \\
RMSProp hyperparameter $\alpha$     & 0.99      \\
RMSProp hyperparameter $\epsilon$     & $1 \times 10^{-5}$     \\
Dimension of the hidden state for agent network     & 64      \\
Latent dimension of an agent     & 16      \\
Dimension of the embedding layer for agent network     & 4      \\
Dimension of the embedding layer for mixing network     & 32      \\
Dimension of the embedding layer for hyper-network     & 64      \\
Number of hyper-network layers     & 2      \\
Runner                                  & Episodic runner  \\
Starting value of exploration rate                & $1$  \\
Finishing value of exploration rate                & $0.05$  \\
Number of time steps for epsilon rate annealing                & $5 \times 10^{4}$  \\
Number of time steps after which the target networks are updated     & 200    \\
Roll-out horizon    & 3      \\
Whether to use KL balancing     & True      \\
KL balancing $\alpha$    & $0.8$      \\
Number of time steps between test intervals        & 10000         \\
Number of episodes for testing every time     & 32      \\
\hline
\end{tabular}
\label{tab:mmsaparams}
\end{table}

\begin{table}[htbp]
\centering
\caption{The generation probabilities of different unit types under the three races: Protoss, Terran, and Zerg.}
\vspace{1em}
\begin{tabular}{|c|c|l|}
\hline
Race             & Unit Type & \multicolumn{1}{c|}{Generation Probability} \\ \hline
\textbf{Protoss} & Stalker   & 0.45                                        \\ \hline
                 & Zealot    & 0.45                                        \\ \hline
                 & Colossus  & 0.1                                         \\ \hline
\textbf{Terran}  & Marauder  & 0.45                                        \\ \hline
                 & Marine    & 0.45                                        \\ \hline
                 & Medivac   & 0.1                                         \\ \hline
\textbf{Zerg}    & Zergling  & 0.45                                        \\ \hline
                 & Hydralisk & 0.45                                        \\ \hline
                 & Baneling  & 0.1                                         \\ \hline
\end{tabular}
\label{tab:smacprobs}
\end{table}

\begin{table}[htbp]
\centering
\caption{The default settings for SMACv2 environments including \textit{terran\_5\_units}, \textit{protoss\_5\_units}, and \textit{zerg\_5\_units}.}
\vspace{1em}
\begin{tabular}{|l | l|}
\hline
\textbf{Description}                                        & \textbf{Value} \\ \hline
Whether agents receive pathing values                & False             \\
within the sight range as part of observations                                                  &             \\[5pt]
Whether to log messages about                & False             \\
observations, states, actions, and rewards     &         \\[5pt]
Whether to use a combination of agents' observations        & False             \\
as the global state                        &             \\[5pt]
Whether agents receive the health of                & True             \\
other agents within the sight range                &             \\[5pt]
Whether to restrict every agent's                & False             \\
field-of-view to a cone                &             \\[5pt]
Whether to use a non-learning heuristic AI        & False             \\[5pt]
Whether agents receive their own health                & True             \\[5pt]
Whether to scale down the reward for every episode      & True             \\[5pt]
Number of time steps              &   5000000 \\[5pt]
Number of allies              &   5 \\[5pt]
Number of enemies             &   5 \\[5pt]
Reward scaling rate                & 20             \\[5pt]
Reward for winning (all enemies die)                     & $+200$         \\[5pt]
Reward for defeating one enemy                     & $+10$        \\[5pt]
Reward when one ally dies                   & $-5$      \\[5pt]
Races                & \{Protoss, Terran, Zerg\}            \\[5pt]
Basic Actions   & \{attack, move, stop, heal, no-op\}            \\[5pt]
Directions         & \{East, West, South, North\}        \\[5pt]
Difficulty                       & 7 (very difficult)     \\[5pt]
Game version                & SC2.4.10            \\[5pt]
\hline
\end{tabular}
\label{tab:smacparams}
\end{table}

\begin{table}[htbp]
\centering
\caption{The default settings for Multi-Agent MuJoCo control suite.}
\vspace{1em}
\begin{tabular}{|l | l|}
\hline
\textbf{Name}                                        & \textbf{Value} \\ \hline

Partitioning of the robot \textit{Ant}    & (hip4, ankle4, hip1, ankle1,  \\
    & hip2, ankle2, hip3, ankle3) \\
Partitioning of the robot \textit{Half Cheetah}    & (bthigh, bshin, bfoot, fthigh, fshin, ffoot) \\
Partitioning of the robot \textit{Hopper}    & (thigh\_joint, leg\_joint, foot\_joint) \\
Partitioning of the robot \textit{Walker2d}    & (foot\_joint, leg\_joint, thigh\_joint,  \\
 & foot\_left\_joint, leg\_left\_joint, thigh\_left\_joint) \\
Number of time steps        &   7000000 \\
Number of nearest joints to observe                    & 2 \\
Batch size                          & 32  \\
Number of episodes for testing during the test interval     & 32 \\
Number of time steps between testing intervals       & 10000  \\
Number of time steps between logging intervals       & 10000  \\         
Whether the agent receives its ID          & True \\
Whether agents receive the last actions &  True \\ 
within the sight range    &  \\
Scaling factor for the returns of \textit{Ant}\_8\texttimes1  & 0.2  \\
Scaling factor for the returns of \textit{HalfCheetah}\_6\texttimes1  & 0.2  \\
Scaling factor for the returns of \textit{Hopper}\_3\texttimes1  & 0.1  \\
Scaling factor for the returns of \textit{Walker2d}\_6\texttimes1  & 0.1  \\

\hline
\end{tabular}
\label{tab:mujocoparams}
\end{table}

\begin{table}[htbp]
\centering
\caption{The default settings for Level-Based Foraging environments.}
\vspace{1em}
\begin{tabular}{|l | l|}
\hline
\textbf{Name}                                        & \textbf{Value} \\ \hline

Number of time steps        &   2000000 \\
Number of episodes for testing during the test interval     & 100 \\
Number of time steps between testing intervals       & 50000  \\
Number of time steps between logging intervals       & 50000  \\ 
Basic Actions   & \{None, North, South, East, West, Load\}            \\
Whether the agent receives its ID          & True \\
Whether agents receive the last actions & True  \\ 
within the sight range    &  \\
\hline
\end{tabular}
\label{tab:lbfparams}
\end{table}

\clearpage
\section{Markov Games}
In MARL environments, the dynamics of the environment and the reward are determined by the joint actions of all agents, and a generalization of MDP that is able to model the decision-making processes of multiple agents is needed. This generalization is known as stochastic games, also referred to as Markov games \cite{Littman1994MarkovGA}. 

\begin{definition}
A Markov game $G_M$ is defined as a tuple $$ \langle \mathcal{N}, \mathcal{S},
\{\mathcal{A}^i \}_{i \in \{1,...,N \}}, P, \{\mathbf{R}^i \}_{i \in \{1,...,N \}}, \mathit{\gamma} \rangle, \text{where}$$
\begin{itemize}
    \item $\mathcal{N} = \{ 1,..., N \}$ represents the set of $N$ agents, and it is equivalent to a single-agent MDP when $N = 1$. 
    \item $\mathcal{S}$ is the state space of all agents in the environment.
    \item $\{\mathcal{A}^i \}_{i \in \{1,...,N \}}$ stands for the set of action space of every agent $i\in \{1,...,N \}$. 
    \item Define $\mathcal{A} := \mathcal{A}^1 \times ... \times \mathcal{A}^N$ to be the set of all possible joint actions of $N$ agents, and define $\triangle{\mathcal{(S)}}$ to be the probability simplex on $\mathcal{S}$. $P : \mathcal{S} \times \mathcal{A} \to \triangle{\mathcal{(S)}}$ is the probability function that outputs the transition probability to any state $s' \in \mathcal{S}$ given state $s \in \mathcal{S}$ and joint actions $\mathbf{a} \in \mathcal{A}$. 
    \item $\{\mathbf{R}^i \}_{i \in \{1,...,N \}}$ is the set of reward functions of all agents, in which $\mathbf{R}^i : \mathcal{S} \times \mathcal{A} \times \mathcal{S} \to \mathbb{R}$ denotes the reward function of the $i$-th agent that outputs a reward value on a transition from state $s \in \mathcal{S}$ to state $s' \in \mathcal{S}$ given the joint actions of all agents $\mathbf{a} \in \mathcal{A}$.
    \item Lastly, $\mathit{\gamma} \in [0,1]$ represents the discount factor w.r.t. time.
\end{itemize}
\end{definition}

A Markov game, when used to model the learning of multiple agents, makes the interactions between agents explicit. At an arbitrary time step $t$ in the Markov game, an agent $i$ takes its action $a_{i,t}$ at the same time as any other agent, given the current state $s_t$. The agents' joint actions, $\mathbf{a}_t$, cause the transition to the next state $s_{t+1} \sim P(\cdot | s_t, \mathbf{a}_t)$ and make the environment to generate a reward $\mathbf{R}^i$ for agent $i$. Every agent has the goal of maximizing its own long-term reward, which can be achieved by finding a behavioral policy
$$\pi(\mathbf{a} | s) := \prod_{i \in \mathcal{N}} \pi^i({a^i} | s).$$

Specifically, the value function of agent $i$ is defined as 
\begin{equation}\label{mg1}
V^i_{\pi^i, \pi^{-i}}(s) :=
\mathbb{E}_{s_{t+1} \sim P(\cdot | s_t, \mathbf{a}_t), a^{-i} \sim \pi^{-i}(\cdot| s_t)}
\big{[} \sum_t \gamma^t \mathbf{R}^i_t(s_t, \mathbf{a}_t, s_{t+1}) | a^{i}_{t} \sim \pi^i(\cdot| s_t), s_0 \big{]} \nonumber
\end{equation}

where the symbol $-i$ stands for the set of all indices in $\{ 1,..., N \}$ excluding $i$. It is clear that in a Markov game, the optimal policy of an arbitrary agent $i$ is always affected by not only its own behaviors but also the policies of the other agents. This situation gives rise to significant disparities in the approach to finding solutions between traditional single-agent settings and multi-agent reinforcement learning.

In the realm of MARL, a prevalent situation arises where agents do not have access to the global environmental state. They are only able to make observations of the state by leveraging an observation function. This scenario is formally defined as Dec-POMDP. It contains two extra terms for the observation function
$O(s, i): \mathcal{S} \times \mathcal{N} \to \Omega$
and for the set of observations $\Omega$ made by each of the agents, in addition to the definition of the Markov game.

\section{Value Decomposition and Individual-Global-Max}
Based on the CTDE paradigm, value decomposition is an effective technique deployed in MARL as it encourages collaboration between agents \cite{son_qtran_2019}. To use this technique, we define the condition of Individual-Global-Max (IGM):
\begin{definition}
Let $\mathcal{A}$ be the joint action space and $\mathcal{T}$ be the joint action-observation history space. Denote the agents' joint action-observation histories as $\boldsymbol{\tau} \in \mathcal{T}$ and their joint actions as $\mathbf{a} \in \mathcal{A}$. Given the joint action-value function $Q_{tot} : \mathcal{T} \times \mathcal{A} \to \mathbb{R},$ if there exist individual $\{ Q_i : \mathcal{T}^{i} \times \mathcal{A}^i \to \mathbb{R} \}_{i \in \{ 1,...,N \}}$ such that the following holds:
\begin{equation}\label{igm1}
\argmax_{\mathbf{a} \in \mathcal{A}} Q_{tot}(\boldsymbol{\tau}, \mathbf{a}) = 
\begin{bmatrix}
    \argmax_{a^1} Q_{1}(\tau^1, a^1) \\
    \vdots \\
    \argmax_{a^N} Q_{N}(\tau^N, a^N) 
\end{bmatrix}
\end{equation}
then $\{ Q_i \}_{i \in \{ 1,...,N \}}$ satisfy the Individual-Global-Max (IGM) condition for $Q_{tot}$ with $\boldsymbol{\tau}$, which means $Q_{tot}(\boldsymbol{\tau}, \mathbf{a})$ can be decomposed by $\{ Q_i \}_{i \in \{ 1,...,N \}}$.
IGM indicates that a MARL task can be solved in a decentralized manner as long as local and global Q-functions are consistent \cite{sunehag_value-decomposition_2017, rashid_qmix_2018, son_qtran_2019}. 
\end{definition}

In order to guarantee that the IGM condition holds, different assumptions have been made. VDN \cite{sunehag_value-decomposition_2017} utilizes a sufficient condition, named \textit{additivity}, for IGM:
$$Q_{tot}(\boldsymbol{\tau}, \mathbf{a}) = \sum^{N}_{i=1} Q_i (\tau^i, a^i)$$
VDN is able to factorize the joint value function assuming the additivity of individual value functions. Alternatively, QMIX \cite{rashid_qmix_2018} uses a sufficient condition called \textit{monotonicity} for IGM and proves that IGM is guaranteed under this assumption:
$$\frac{\partial Q_{tot}(\boldsymbol{\tau}, \mathbf{a})}{\partial Q_i (\tau^i, a^i)} \geq 0, \forall i \in \{ 1,...,N \}.$$ 
Lastly, QTRAN \cite{son_qtran_2019} proposes to find individual action-value functions [$Q_i$] that factorize the original joint action-value function $Q_{jt}$ and to transform $Q_{jt}$ into a new value function $Q_{jt}'$ that shares the same optimal joint action with $Q_{jt}$. The sufficient condition for [$Q_i$] to satisfy the IGM is described in detail in Theorem 1 of \cite{son_qtran_2019}.

\section{Structured Variational Inference}

The process of structured variational inference involves approximating some complicated distribution $p(\mathbf{y})$ with $q(\mathbf{y})$, another potentially simpler distribution \cite{Levine-1805-00909}. The approximate inference is performed by optimizing the variational lower bound. \cite{Huang2020SVQNSV} introduces a probabilistic graphical model (PGM) for single-agent partially observable settings and solves the variational inference problem under the model. The variational lower bound associated with the PGM can be written as follows:
\begin{align*}
\log p(\mathcal{O}_{0:T}, {a}_{0:T}, {o}_{1:T}) \nonumber
&= \log \mathbb{E}_{q_{\theta}({s}_{1:T} | \mathcal{O}_{1:T}, {a}_{0:T},{o}_{1:T} )} 
\left[ \frac{p({s}_{1:T}, \mathcal{O}_{0:T}, {a}_{0:T}, {o}_{1:T} )}{q_{\theta}({s}_{1:T} | \mathcal{O}_{0:T}, {a}_{0:T}, {o}_{1:T} )} \right]  \nonumber \\
& \geq \mathbb{E}_{q_{\theta}({s}_{1:T} | \mathcal{O}_{1:T}, {a}_{0:T},{o}_{1:T})} 
\log \left[ \frac{p({s}_{1:T}, \mathcal{O}_{0:T}, {a}_{0:T}, {o}_{1:T} )}{q_{\theta}({s}_{1:T} | \mathcal{O}_{0:T}, {a}_{0:T}, {o}_{1:T} )} \right],  \nonumber \\
\end{align*}
where $s, a, o$ are the state, action, and observation of the agent, $q_{\theta}$ is the approximate function, and $\theta$ stands for the learnable parameter. $\mathcal{O}_t$ is a binary random variable introduced by \cite{Levine-1805-00909} related to maximum entropy reinforcement learning. It indicates the optimality of the action at time $t$.

\newpage
\section{Detailed Deduction of the ELBO}
\label{appx:elbo}

Equation \ref{mtd1} in the main paper briefly shows how the ELBO of Dec-POMDP, 
$\mathcal{L}_{\text{ELBO}} (\mathbf{a}_{0:T}, \mathbf{o}_{1:T})$,
is derived. We present the full deduction process here.

\begin{align}
&\quad \mathcal{L}_{\text{ELBO}} (\mathbf{a}_{0:T}, \mathbf{o}_{1:T}) \nonumber\\
&=\log p\left( \mathbf{a}_{0:T}, \mathbf{o}_{1:T} \right) \nonumber\\
&= \log \mathbb{E}_{q_{\theta}(\hat{s}_{1:T} \mid \mathbf{a}_{0:T}, \mathbf{o}_{1:T} )} 
\left[ \frac{p(\hat{s}_{1:T}, \mathbf{a}_{0:T}, \mathbf{o}_{1:T} )}{q_{\theta}(\hat{s}_{1:T} \mid \mathbf{a}_{0:T}, \mathbf{o}_{1:T} )} \right]  \nonumber \\
& \geq \mathbb{E}_{q_{\theta}(\hat{s}_{1:T} \mid \mathbf{a}_{0:T}, \mathbf{o}_{1:T} )} 
\log \left[ \frac{p(\hat{s}_{1:T}, \mathbf{a}_{0:T}, \mathbf{o}_{1:T} )}{q_{\theta}(\hat{s}_{1:T} \mid \mathbf{a}_{0:T}, \mathbf{o}_{1:T} )} \right] \label{elboapd1} \\
&=\int q_{\theta}(\hat{s}_{1:T} \mid \mathbf{a}_{0:T}, \mathbf{o}_{1:T} ) \log \left[ \frac{p(\hat{s}_{1:T}, \mathbf{a}_{0:T}, \mathbf{o}_{1:T} )}{q_{\theta}(\hat{s}_{1:T} \mid \mathbf{a}_{0:T}, \mathbf{o}_{1:T} )} \right] d \hat{s}_{1:T} \nonumber \\
&=\int \sum_{t=1}^{T} q_{\theta}(\hat{s}_{1:T} \mid \mathbf{a}_{0:T}, \mathbf{o}_{1:T} ) \log \left[\frac{ p\left(\mathbf{a}_t \mid \mathbf{o}_{t}\right) p\left(\mathbf{o}_{t} \mid \hat{s}_{t}\right) p\left(\hat{s}_{t} \mid \hat{s}_{t-1}, \mathbf{a}_{t-1}\right) }{q_{\theta}\left(\hat{s}_{t} \mid \hat{s}_{t-1}, \mathbf{a}_{t-1}, \mathbf{o}_{t}\right)}\right] d \hat{s}_{1:T} \nonumber\\
&=\sum_{t=1}^{T} \int q_{\theta}(\hat{s}_{1:t} \mid \mathbf{a}_{0:t}, \mathbf{o}_{1:t} ) \log \left[\frac{ p\left(\mathbf{a}_t \mid \mathbf{o}_{t}\right) p\left(\mathbf{o}_{t} \mid \hat{s}_{t}\right) p\left(\hat{s}_{t} \mid \hat{s}_{t-1}, \mathbf{a}_{t-1}\right) }{q_{\theta}\left(\hat{s}_{t} \mid \hat{s}_{t-1}, \mathbf{a}_{t-1}, \mathbf{o}_{t}\right)}\right] d \hat{s}_{1:t} \nonumber\\
&=\sum_{t=1}^{T}\left\{\int q_{\theta}(\hat{s}_{1:t} \mid \mathbf{a}_{0:t}, \mathbf{o}_{1:t} ) \log \left[ p\left(\mathbf{a}_t \mid \mathbf{o}_{t}\right) p\left(\mathbf{o}_{t} \mid \hat{s}_{t}\right)\right] d \hat{s}_{1:t} \right. \nonumber\\
&\quad \left.+\int q_{\theta}(\hat{s}_{1:t} \mid \mathbf{a}_{0:t}, \mathbf{o}_{1:t} ) \log \left[\frac{p\left(\hat{s}_{t} \mid \hat{s}_{t-1}, \mathbf{a}_{t-1}\right)}{q_{\theta}\left(\hat{s}_{t} \mid \hat{s}_{t-1}, \mathbf{a}_{t-1}, \mathbf{o}_{t}\right)}\right] d \hat{s}_{1:t}\right\}\nonumber\\
&=\sum_{t=1}^{T}\left\{\int q_{\theta}(\hat{s}_{1:t} \mid \mathbf{a}_{0:t}, \mathbf{o}_{1:t} ) \log \left[ p\left(\mathbf{a}_t \mid \mathbf{o}_{t}\right) p\left(\mathbf{o}_{t} \mid \hat{s}_{t}\right)\right] d \hat{s}_{1:t} \right. \nonumber\\
&\quad \left.-\int q_{\theta} \left(\hat{s}_{1:t-1} \mid \mathbf{a}_{0:t-1}, \mathbf{o}_{1:t-1} \right) \mathcal{D}_{\text{KL}} \left[q_{\theta}\left(\hat{s}_{t} \mid \hat{s}_{t-1}, \mathbf{a}_{t-1}, \mathbf{o}_{t}\right) \parallel p\left(\hat{s}_{t} \mid \hat{s}_{t-1}, \mathbf{a}_{t-1}\right)\right] d \hat{s}_{1:t}\right\}\nonumber\\
&= \mathbb{E}_{q_{\theta} \left(\hat{s}_{1:T} \mid \mathbf{a}_{0:T}, \mathbf{o}_{1:T} \right)} \sum_{t=1}^{T} \left\{ \log \left[ p\left(\mathbf{a}_t \mid \mathbf{o}_{t}\right) p\left(\mathbf{o}_{t} \mid \hat{s}_{t}\right)\right] \right.\nonumber\\ 
&\left.\quad- \mathcal{D}_{\text{KL}} \left[q_{\theta}\left(\hat{s}_{t} \mid \hat{s}_{t-1}, \mathbf{a}_{t-1}, \mathbf{o}_{t}\right) \parallel p\left(\hat{s}_{t} \mid \hat{s}_{t-1}, \mathbf{a}_{t-1}\right)\right] \right\} \nonumber\\ 
&\approx \sum_{t=1}^{T} \{\log \left[ p\left(\mathbf{a}_t \mid \mathbf{o}_{t}\right) p\left(\mathbf{o}_{t} \mid \hat{s}_{t}\right)\right] \nonumber\\ 
&\quad - \mathcal{D}_{\text{KL}} \left[q_{\theta}\left(\hat{s}_{t} \mid \hat{s}_{t-1}, \mathbf{a}_{t-1}, \mathbf{o}_{t}\right) \parallel p\left(\hat{s}_{t} \mid \hat{s}_{t-1}, \mathbf{a}_{t-1}\right)\right]\} \label{elboapd2}\\
&= \sum_{t=1}^{T} \{\log \left[ p\left(\mathbf{a}_t \mid \mathbf{o}_{t}\right) \right] + \log \left[ p\left(\mathbf{o}_{t} \mid \hat{s}_{t}\right)\right] \nonumber\\ 
&\quad - \mathcal{D}_{\text{KL}} \left[q_{\theta}\left(\hat{s}_{t} \mid \hat{s}_{t-1}, \mathbf{a}_{t-1}, \mathbf{o}_{t}\right) \parallel p\left(\hat{s}_{t} \mid \hat{s}_{t-1}, \mathbf{a}_{t-1}\right)\right]\} \label{elboapd3}
\end{align}

Note that (\ref{elboapd1}) is reached via Jensen's inequality. Equation \ref{elboapd2} can be obtained by sampling 
$\hat{s}_{1:T} \sim q_{\theta}\left(\hat{s}_{1:T} \mid \mathbf{a}_{0:T}, \mathbf{o}_{1:T} \right)$. Equation \ref{elboapd3} is identical to Equation \ref{mtd1}.

\section{Limitations and Broader Impact}

Our research primarily emphasizes the practical application and evaluation of the proposed MARL approach. Our focus on experiments in various benchmarks is intentional, as we aim to empirically demonstrate the generalizability and robustness of our method. While we recognize the significance of theoretical guarantees in MARL, the complexities of multi-agent systems and our method present great challenges in deriving rigorous mathematical proofs. Addressing the theoretical side of joint state-action representation learning within the MARL framework could be a focus of our future work.

Our work aims at contributing to the development of multi-agent and model-based RL algorithms. Although this development could be applied in various fields, it does not require specific ethical considerations to be highlighted. Nonetheless, it is crucial to emphasize the importance of responsible deployment to ensure a beneficial impact on society.

\end{document}